\documentclass[preprint,12pt]{elsarticle}

\usepackage{lineno,hyperref}
\usepackage{diagbox}
\usepackage{amsmath}
\usepackage{amsfonts}
\modulolinenumbers[5]
\usepackage[colorinlistoftodos]{todonotes}

\usepackage{subcaption}
\usepackage{xcolor}

\journal{arXiv}
\begin{document}

\begin{frontmatter}

\title{Traffic Modelling and Prediction via Symbolic Regression on Road Sensor Data}

\author{Alina Patelli, Victoria Lush, Anik\'o Ek\'art}
\address{Aston Lab for Intelligent Collectives Engineering (ALICE), Aston University}
\author{Elisabeth Ilie-Zudor}
\address{Institute for Computer Science and Control, Hungarian Academy of Sciences}



%\author{
%\IEEEauthorblockN{
%Alina Patelli, Victoria Lush, Anik\'o Ek\'art and Elisabeth Ilie-Zudor 
%}\\
%\IEEEauthorblockA{
%\begin{tabular}{cc}
% \begin{tabular}{@{}c@{}}
%\Mark{1} Aston Lab for Intelligent \\Collectives Engineering (ALICE) \\ Aston University, Birmingham, UK \\\{a.patelli2, v.lush1, a.ekart\}@aston.ac.uk}
%\end{tabular} &
%\begin{tabular}{@{}c@{}}
%\Mark{2} 
%Institute for Computer Science and Control\\Hungarian Academy of Sciences\\ Budapest, Hungary \\
%ilie@sztaki.mta.hu
%\end{tabular}
%\end{tabular}
%}
%\thanks{A. Patelli, V. Lush and A. Ek\'art are with Aston Lab for Intelligent Collectives Engineering (ALICE), Aston University, Birmingham, UK, \{a.patelli2, v.lush1, a.ekart\}@aston.ac.uk. E. Ilie-Zudor is with the  Institute for Computer Science and Control, Hungarian Academy of Sciences, Budapest, Hungary, ilie@sztaki.mta.hu}
%\thanks{This work was supported by the European Commission through the H2020 project EXCELL (http://excell-project.eu/), grant No. 691829.}
%}

\begin{abstract}
The continuous expansion of the urban traffic sensing infrastructure has led to a surge in the volume of widely available road related data. 
Consequently, increasing effort is being dedicated to the creation of intelligent transportation systems, where decisions on issues ranging from city-wide road maintenance planning to improving the commuting experience are informed by computational models of urban traffic instead of being left entirely to humans. The automation of traffic management has received substantial attention from the research community, however, most approaches target highways, produce predictions valid for a limited time window or require expensive retraining of available models in order to accurately forecast traffic at a new location. In this article, we propose a novel and accurate traffic flow prediction method based on symbolic regression enhanced with a lag operator. Our approach produces robust models suitable for the intricacies of urban roads, much more difficult to predict than highways. Additionally, there is no need to retrain the model for a period of up to 9 weeks. Furthermore, the proposed method generates models that are transferable to other segments of the road network, similar to, yet geographically distinct from the ones they were initially trained on. We demonstrate the achievement of these claims by conducting extensive experiments on data collected from the Darmstadt urban infrastructure.

%Journal: Transportation Research Part C: Emerging Technologies 3.805

%Alternative: IEEE Trans on Intelligent Transport IF 3.724

\end{abstract}

\begin{keyword}
traffic prediction \sep traffic modelling \sep genetic programming \sep symbolic regression
%\texttt{elsarticle.cls}\sep \LaTeX\sep Elsevier \sep template
%\MSC[2010] 00-01\sep  99-00
\end{keyword}

%\IEEEpeerreviewmaketitle

\end{frontmatter}

%\linenumbers
\section{Introduction}

%From workforce mobility to supply chain logistics, crucial aspects of modern society rely on road transportation. 
%\todo{Any public source to cite that gives percentage/volume of transport on road, train, water, air? Anything about time/distance people commute to work?}
In 2014, 83.4\% of all inland passenger transport across EU states was accomplished by car, 9.1\% by coaches, buses and trolleys, leaving only 7.6\% to cleaner alternatives such as train travel.\footnote{\url{http://ec.europa.eu/eurostat/statistics-explained/index.php/Passenger_transport_statistics\#Main_statistical_findings}} European freight transportation mainly occurs on roads, namely 74.9\%, leaving only a share of 18.4\% to be fulfilled on rail and 6.7\% on inland waterways.\footnote{\url{http://ec.europa.eu/eurostat/statistics-explained/index.php/Freight_transport_statistics\#Modal_split}} Furthermore, in 2015, there were 25 regions within the European Union where over one fifth of the workforce commuted to work.\footnote{\url{http://ec.europa.eu/eurostat/statistics-explained/index.php/Statistics_on_commuting_patterns_at_regional_level\#National_commuting_patterns}} These European Commission statistics provide strong evidence in support of the fact that industrialisation and overpopulation have pushed urban infrastructures to and beyond capacity~\cite{xu2014}, stemming severe problems around traffic safety, road maintenance and development costs as well as air and noise pollution, to name only a few. The smart cities vision \cite{camero2019smart,chourabi2012,liu2017exploring,ahvenniemi2017differences,zhuhadar2017next,diao18} offers a potential solution to all these issues, in the form of Intelligent Transport Systems (ITS) \cite{banaei2011,sjoberg2017cooperative,mangiaracina2017comprehensive,baumler2017intelligent,kalamaras2018interactive,zhu2018big}. These typically span over a number of connected junctions, are predicated on automated traffic control and entail three stages:
\begin{itemize}
\item \textit{Monitoring} intersections and collecting data, such as traffic volume and speed, using a variety of road sensors (cameras, induction loops, etc.).
\item \textit{Modelling} traffic through intersections by expressing outflow data as a function of inflow data. 
\item \textit{Prediction} of future traffic patterns for decision support, both real-time and long term, including assessing the length of a commute, estimating the pollution level on a cycle route and advising in which road segments the city council should invest, to absorb increasing traffic density.
\end{itemize} 

In the age of big data, the practical implementation of ITS is approaching fruition at an unprecedented speed. Following a wide expansion of the under-, at- and over-road level sensor arrays across major cities, impressively large volumes of traffic related readings are fast becoming available. There is a noticeable trend across Europe (e.g., Birmingham,\footnote{\url{http://dmtlab.bcu.ac.uk/alandolhasz/birt/index.html}} Darmstadt\footnote{\url{https://darmstadt.ui-traffic.de/faces/TrafficData.xhtml}}), at city council level, to invest in installing and expanding traffic sensing infrastructures as well as making the resulting data publicly available. However, the practical usability of this large amount of data is undercut by two significant issues. Firstly, accurate and reliable traffic readings are difficult to collect due to a variety of causes, such as malfunctioning or improperly installed sensors, sub-optimally configured servers requiring specialised scripts to download the traffic readings they host, etc. Secondly, transportation data are not actionable in  raw form \cite{li2015robust}  - indeed, car counts and average speed values  catalogued in endless spreadsheets would do little to assist commuters in selecting the shortest or fastest way to work.  Hence, a sensible implementation of ITS would require an automated framework capable of (1) tackling the collection and actionability issues surrounding road traffic data, (2) producing a relevant (timely as well as accurate \cite{lv2015}) prediction and (3) presenting it to the stakeholders in an intuitive, human-readable form, that supports ``pro-active decision making'' \cite{smith1994short}. 

The research community has made significant efforts towards achieving such a sensible implementation of ITS, by employing a wide range of modelling and prediction approaches. Here, we give an overview of the progress made with respect to each of the three ITS stages. %- following this, section 2 [replace with section ref] will provide details of the methods and achievements.
%\todo{Perhaps, this statement about what is to follow can go in the last paragraph of the Intro}

\paragraph*{Monitoring} There is a noticeable gap in the literature with respect to the efficient collection of large volumes of accurate traffic data, which is an absolutely necessary precursor of modelling and prediction. Most researchers report on the size, type and sampling frequency \cite{jeon2016monte,stathopoulos2003multivariate,min2011real} of the sensor array furnishing data but exclude any mention of how to address problems such as missing or corrupted samples, noise, outliers in the data stream, miss-labelled records  and absent contextual information (e.g., sports events, road works, congestion etc., at the time of data collection). Real data are rarely 100\% accurate and complete \cite{meeker2014reliability}, hence, without preprocessing, the collected data by themselves are not sufficient for reliable and robust modelling and prediction. We note that \textit{an ITS will only be accepted by the general public if the offered decision support is of use.} 

\paragraph*{Modelling} Most research concentrates on very frequent
% is "sharp" a good adjective?
sampling, going as far as to develop models based on data collected every 30 seconds. This is necessary for emergency response planning but shows limited applicability in medium and long term problems,  such as estimating daily commute times or informing a five year road infrastructure extension strategy. Furthermore, a substantial proportion of contributions study freeway traffic \cite{banaei2011,li2015robust,shi2015big}, which is significantly simpler to analyse than intra-city junctions, where the physical layout is much more complex and the daily/weekly patterns are significantly less predictable. The reviewed literature overwhelmingly investigates regression-based models \cite{jeon2016monte,stathopoulos2003multivariate,min2011real}, neural networks - deep learning approaches \cite{zeng2013development} or combinations of the above \cite{quek2006pop,lv2015}. The first category comprises AR(I)MA(X) models, structurally simple, therefore fairly inexpensive to synthesise, yet not sufficiently accurate, except in specifically defined cases. The second category generally boasts more accurate, however implicitly computationally expensive models, which are mostly difficult to re-use on road configurations other than those used for training \cite{smith1994short}. In addition, traffic abstractions derived via deep learning or one of its variants are opaque \cite{smith1994short,shi2015big}: they are overly complex, with many parameters and the physical significance of such models (i.e., mapping model parameters to the specific junction arms 
% Aniko: is "arm" the correct term? Road segment may be better, but leave for now.
they quantify) is difficult to infer. \textit{There is a need for modelling traffic on larger time horizons to capture the characteristics needed for medium and long term problems.}

\paragraph*{Prediction} Most models derived in the previous stage are utilised, with acceptable accuracy, for short term predictions, usually in the same location where the model was obtained \cite{smith1994short,jeon2016monte,stathopoulos2003multivariate}. Thus, \textit{estimating traffic patterns for weeks or longer in the future and in structurally similar yet different areas of the road network has not yet been explored, to our knowledge.}   

\vspace{10pt}

To address the shortcomings identified above, we propose an ITS based on symbolic regression, an evolutionary algorithm drawing on genetic programming \cite{koza1992}. As we will demonstrate in the remainder of the paper, our approach tackles the issues related to the three stages of ITS in the manner outlined below.

\paragraph*{Monitoring} Symbolic regression models are robust enough to tolerate missing samples in their training data, without significant compromises in accuracy. Consequently, our \textbf{claim 1} is that \textbf{our approach features high tolerance to sensor faults}.

\paragraph*{Modelling} The accuracy of approximations with symbolic regression improves for longer sampling intervals. More specifically, the approximation errors of symbolic regression models on both training and test data are increasing at a lower rate compared to data sampling interval lengths (e.g., as the training data collection window is extended from 5 to 15 minutes, the testing error slightly decreases, as shown in Table \ref{tabSRTrainOnAllTestOn15}). Moreover, our models are built for intra-city (urban) traffic, as opposed to highways, and are easily interpretable, in the sense that each model term can be clearly correlated with a specific junction arm.
% arm?
In addition, symbolic regression inherently eliminates irrelevant input without compromising the accuracy of the model's output (the junction arm 
% arm?
that is being modelled). This leads to our \textbf{claim 2}, in that \textbf{our models are robust and reliable} (a), irrespective of the training window, \textbf{flexible} (b) enough to model complex urban traffic layout and dynamics (spatio-temporal  feasibility) and \textbf{self-managing} (c), in terms of eliminating irrelevant terms. 

\paragraph*{Prediction} The traffic patterns we estimate based on symbolic regression models are long-term (the accuracy of the prediction remains satisfactory when using a model trained up to 9 weeks in the past) and applicable to structurally similar yet different junctions. Our \textbf{claim 3} is that \textbf{our predictions have a long shelf-life} (a), that is, they remain valid for an extended period of time, without the need to retrain the model, \textbf{and are easy to understand, suitable for effective and confident decision making} (b). Specifically, our approach could be used to answer infrastructure-relevant questions concerning, for instance, the financial feasibility of installing a sensor at a currently unmonitored location. Should a sufficiently accurate prediction of traffic through that node be obtained with one of our models trained on a similar junction in a different part of the city, investing in hardware for the new location may become unnecessary.
%transfer learning link comes here
The generalisation of a model that was previously trained on similar, but not identical data falls under the umbrella of transfer learning \cite{pan2010survey}, more precisely transductive transfer learning \cite{arnold2007transtl}. Investigations on the potential of GP, and in particular symbolic regression, for transfer learning are already under way and the results on specifically designed benchmarks for three types of transfer are promising \cite{haslam2016gptl}.

After reviewing the state of the art contributions with respect to the three stages of ITS (section \ref{secBackground}), we present the Darmstadt city traffic data, as retrieved from the Urban Institute's web portal.\footnote{https://darmstadt.ui-traffic.de/faces/TrafficData.xhtml} Section \ref{secStateOfTheArt} offers a summary of the state of the art methods for predicting traffic, followed by a detailed presentation of the specialised symbolic regression approach we propose (section \ref{secSymbRegProposed}). The experimental results supporting our claims are the topic of section \ref{secExpRes}, whereas the final  part of the paper is dedicated to conclusions and future work.

\section{Background}
\label{secBackground}
%There is a significant body of work dedicated to applications of evolutionary approaches to practical problems: EC tools to all problem categories - Soft Computing and other journals
%\subsection{Transportation Problems}
%fairly standard EC tools applied to transportation problems
%\subsection{Genetic Programming}
%on all problems and transportation ones

 The research community has proposed a wide variety of solutions to address the three ITS stages, with an overwhelming focus on the final two. The most striking commonality that transpires from analysing the state of the art with respect to road traffic \textit{modelling} and \textit{prediction} is the acknowledgement and exploitation of the spatio-temporal aspect. The spatial component of road traffic refers to the influence exerted on a specific point (e.g., junction) by vehicle flow through connected arteries, both upstream and downstream from the considered point. The temporal facet describes the consistency of traffic patterns applicable to a well-defined time window: for instance, morning traffic does not differ substantially across the work week, unless there is an accident or road maintenance. This is useful when modelling and predicting traffic, as illustrated in the contributions reviewed in the following.
 
\subsection{Monitoring}
The finer details of the first ITS component are given little attention in the reviewed body of work. We note that the most popular sources of traffic data are the California freeway system \cite{shi2015big,li2015robust,banaei2011} and the Far East - China \cite{castro2012urban}, Japan \cite{chen2016learning} and Singapore \cite{quek2006pop}. Traffic data obtained in the traditional way, from on-road sensors (cameras, induction loops) and in-car GPS, are sometimes combined with relevant information extracted from social media platforms \cite{ni2014using}, in the hope of increasing model quality around major sporting events, road works, accidents, etc., that are likely to get extensive coverage on Twitter, Facebook or similar.

\paragraph*{Summary} Firstly, collecting traffic data from highways, rather than urban roads, streamlines \textit{modelling} and \textit{prediction}, as the pronounced seasonality of outer-city traffic is easier to process than the intricacies of a spaghetti junction. Secondly, the transparency of data collection and pre-processing need to be improved - especially with respect to the handling of missing and/or corrupted samples. The literature deals with this aspect only in its most basic form, such as using vicinity averages to replace missing data points \cite{li2015robust}. Finally, road traffic models to be developed in the second ITS stage should show high tolerance to missing data, that is, be robust enough to generate good quality predictions when trained on incomplete data sets.

\subsection{Modelling and Prediction} The approaches relevant to these ITS phases can be divided in three categories \cite{huang2014deep,banaei2011}: time series (parametric models, of which the most popular is the AutoRegressive Integrated Moving Average with eXogenous inputs - AR(I)MA(X) - family \cite{box2015}), probabilistic (Markov chains, Bayesian networks, Monte Carlo simulations) and non-parametric (k-nearest neighbour, artificial neural networks and deep learning, state vector machines).

\subsubsection{Time Series Models}
Determining the parameters of AR(I)MA(X) models is, in most cases, a computationally straightforward problem \cite{maronna2006robust}. Unfortunately, the accuracy obtained with these traditional prediction methods is not generally satisfactory - AR(I)MA(X) models showcase known issues such as inability to model non-stationary anomalies (accidents, road works), low tolerance to missing data in the training set and limited applicability to light nighttime and weekend traffic. However, researchers report promising results when calibrating the classic time series approach to incorporate contextual information, specific to the area/ time window being modelled. For instance, vehicle speed is successfully predicted via time series analyses, after calculating correlation coefficients and applying  Monte Carlo simulation to select the most relevant data samples for training \cite{jeon2016monte}. Another variant proposes multi-variate time series state space models, using data from central Athens, which are produced to map downstream data against upstream data, relative to a specific point on a given corridor \cite{stathopoulos2003multivariate}. The approach produces short term predictions only. A more robust time series model \cite{min2011real} takes into account historical data from junctions neighbouring the one being modelled and produces predictions for several 5 minute intervals in the future. Min and Wynter's approach \cite{min2011real} leverages traffic seasonality - specifically, the number of model parameters is reduced by manually separating training data in peak and off-peak slots. Another possibility is to consider multiple dependent time series, produced by applying the Granger test \cite{li2015robust} (Granger causality is used to measure the degree to which traffic through neighbouring junctions influences vehicle flow through the modelled junction). The daily trend reflected by the dependent time series is then separated from non-stationary bursts (accidents and other anomalies). Finally, Least Absolute Shrinkage and Selection Operator (Lasso) regression is applied to further remove irrelevant data and detect strong dependencies amongst remaining samples. This last step is aimed at parsimony control.

\subsubsection{Probabilistic Models} 
As opposed to time series based approaches, probabilistic alternatives are predicated on computationally expensive simulations. The trade-off is that their general quality is superior to that of AR(I)MA(X) models, hence their popularity in road traffic prediction, either as stand-alone procedures or as precursors to time series analysis. To illustrate the latter category, Monte Carlo simulation is used to select the most relevant road speed data samples \cite{jeon2016monte}, after anomalous readings (collected at the time of major road works, sporting events, etc.) had been eliminated via correlation analysis. On the other hand, a stand-alone probabilistic method is employed for real time crash prediction \cite{shi2015big} by running a Bayesian inference algorithm, following random forest mining. This is done to select features relevant to good quality road traffic prediction, which were revealed to be the logarithm of vehicle volume, the peak hour interval, the average vehicle speed and the congestion index (calculated as the difference between actual road speed and free flow speed, divided by the latter). An alternative is to make use of Markov chains to organise taxi GPS data in transition matrices, to support predicting future traffic flow \cite{castro2012urban}. 

\subsubsection{Non-parametric Models}
Motivated, partly, by their superior local accuracy (once properly trained) as well as by the significant traction gained by deep learning in both scientific circles and popular culture, non-parametric models are well represented in the area of traffic modelling and control. Their main disadvantage that most relevant research is attempting to overcome is opaqueness: models produced by Artificial Neural Networks (ANN) feature a large number of parameters with no apparent links to road traffic features. To address this, one contribution employs time delayed state space neural networks \cite{zeng2013development}, with experimentally selected input. The study revealed that speed or occupancy are sufficient, on their own, for acceptable predictions of freeway traffic. However, a combination of speed, occupancy and volume is necessary for high quality predictions. The spatial aspect is accounted for by using data from both upstream and downstream segments in relation to the one being modelled.

Fuzzy logic is used in combination with a neural network for short term traffic flow prediction \cite{quek2006pop}. The fuzzy layer encodes traffic characteristics that are valuable to traffic engineers (vehicle classification counts and speed) and thus compensate for the opaqueness of the neural network. A deep learning approach based on a stack of auto-encoders (neural networks that attempt to reproduce their input), topped with a prediction layer \cite{lv2015} is deployed to predict traffic on Californian freeways. Several time windows (from 1 to 12 lags) are considered for training  - the number of hidden layers and their size is determined based on the best accuracy achieved on each window. The model is shown to behave well on medium and heavy traffic, but poorly on light traffic. A similar approach entails a stack denoise autoencoder, used to classify features from human mobility data (GPS locations), in order to infer traffic accident rates \cite{chen2016learning}. It is experimentally proven to outperform logistic regression and support vector machine (SVM) alternatives. Huang et al. deploy a deep belief network with a multitask regression layer on top for unsupervised feature learning \cite{huang2014deep}. A task is defined as predicting traffic flow for one road - hence, multitasking is a form of taking into account spatial dependencies. Another contribution in this category \cite{banaei2011} considers the spatial and temporal correlations inherently. A stacked autoencoder model is employed to learn generic traffic flow features after having been trained in a greedy layerwise fashion. The initial set of features are selected manually: length, direction, capacity, connectivity, density, and locality are considered for each road segment taken into account. The stacked autoencoder consequently extracts a subset of features, namely those most closely correlated to the signature traffic flow patterns for the analysed segments. The study concludes that road segments can be partitioned into a set of distinct subpartitions with similar traffic flow patterns and that the combination of direction, connectivity and locality of a road segment can best predict the traffic signature of the segment.

Evolutionary algorithms are efficiently employed to process real-time traffic data and communicate re-routing suggestions to smart spots at traffic lights or LED panels along the road, thus directing drivers to alternative, faster routes \cite{stolfi2019sustainable}. In a similar vein, the Flow Generation Algorithm \cite{stolfi2018generating}, with an evolutionary foundation, consults city topologies available on OpenStreetMap and feeds sensor-collected vehicular data into a traffic simulator to produce realistic approximations of vehicle density at predetermined measurement points. A similar technology is embodied by Green Swarm \cite{stolfi2018green}, an evolutionary mobility architecture which, besides travel time and fuel consumption reduction, also targets the decrease of greenhouse gas emissions. 

\subsubsection{Hybrid Models}
Given that the previously discussed categories of approaches are not universally likely to generate high quality predictions, the research community has created scope for hybridising methods from different groups. To that effect, a simple Bayesian Classifier and a support vector machine are dynamically alternated, based on live traffic conditions, to produce the optimal vehicle flow model \cite{xu2014}. For this purpose, a ``regret metric'' is used to penalise under-performing predictors. The approach is shown to outperform its two constituents, when they are deployed individually. Another attempt to combine optimisation approaches employs ARIMA to calculate the linear component of the traffic model and Genetic Programming (GP) \cite{koza1992} to account for the nonlinear component \cite{xu2016}. After the two are aggregated, the result is shown to outperform stand-alone ARIMA, under both normal and accident conditions. De Souza et al. hybridise a GP component with a boosting technique for time series forecasting \cite{deSouza2009}. The weights involved in the GP base learner's fitness function are updated by the boosting algorithm, based on the correlation between the current traffic prediction and the real data. Traffic light control is improved by using, in turn, two GA variants, Differential Evolution and a multi-objective variant of Evolutionary Decomposition \cite{peres2018multiobjective}, with a reported reduction in travel time and pollution on the streets of Montevideo. An alternative solution to the traffic light programming problem employs Differential Evolution, GA and Particle Swarm Optimisation to leverage simulated traffic scenarios reflective of recorded traffic flow as well as the specific road network topology of the city of Malaga \cite{ferrer2019reliable}. Bederina and Hifi \cite{bederina2018hybrid} approach the vehicle routing problem from a novel angle, targeting not only the reduction of the number of vehicles but also that of travel time. This is achieved by hybridising a multi-objective GA with a ``destroy-rebuild'' technique that encourages solution diversity.
 
\paragraph*{Summary} The most commonly mentioned shortcoming of state of the art traffic modelling and prediction approaches is the computational cost - approximation accuracy trade-off. In the case of ANNs, input (feature) selection is done mostly manually \cite{zeng2013development,banaei2011}, by traffic experts, whereas the structure and size of the  networks' hidden layer is usually configured by trial and error \cite{lv2015, li2015robust}. Automating these operations \cite{shi2015big} ultimately increases the methods' overall complexity even more. Another challenge is selecting the size of the training time window \cite{zeng2013development}, calling for a solution to automatically configure the optimum time lag. The reverse of this problem, namely accurately predicting traffic for a sufficiently large time window in the future, is a problem in and of itself. %The few papers that target urban traffic \cite{stathopoulos2003multivariate} provide short term predictions for single corridors only. 
Authors who target urban traffic, \cite{stathopoulos2003multivariate, vlahogianni2014short, lana2018road}, provide short term predictions for single corridors only, with some evidence of applicability of the obtained models to other corridors.
This brings about the issue of time and space-wise locality of most state of the art solutions - ideally, traffic predictions should h ,old for both heavy (morning and evening) and light (midday and weekend) traffic as well as be applicable on a variety of structurally similar yet geographically different locations, without retraining. Finally, neither probabilistic methods nor non-parametric ones are capable of providing computationally affordable, parsimonious models, with a clear link between featured parameters and modelled traffic features. 

%[Once we settle on the claims in the introduction, I think we should map those against the issues in the state of the art.]

\section{The Darmstadt Case Study}

\begin{figure*}[ht]
\begin{center}
\includegraphics[width=0.9\textwidth]{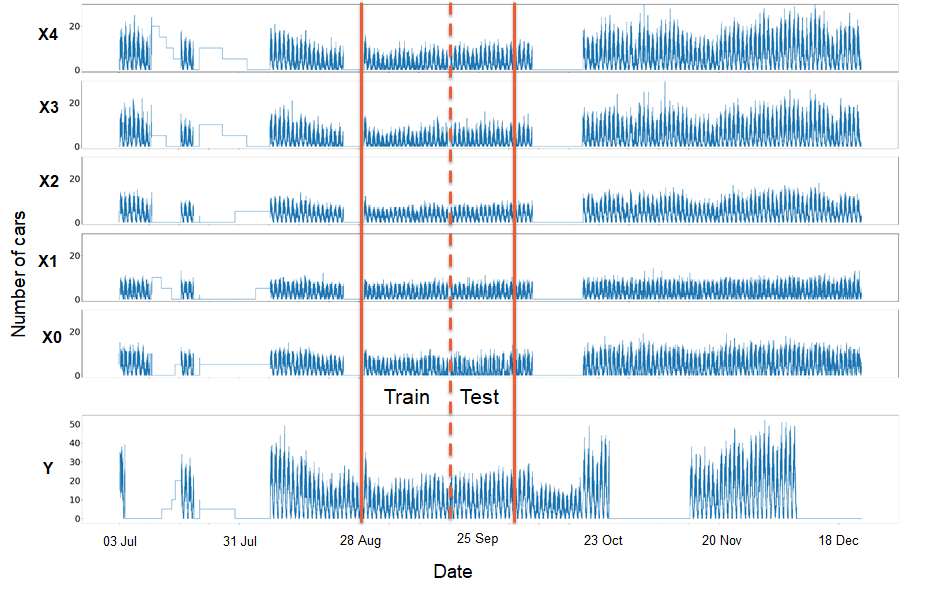}
\end{center}
\caption{Traffic flow for all input and output for junction A13: red lines indicate data used for most experiments - 3 weeks for training (28/08/2017 0:00 - 17/09/2017 23:59); 2 weeks for testing (18/09/2017 0:00 - 01/10/2017 23:59)}
\label{figOverview}
\end{figure*}

Data for this study (see Fig. \ref{figOverview}) were obtained from the Urban Institute\footnote{https://darmstadt.ui-traffic.de/faces/TrafficData.xhtml} (UI) of Darmstadt that hosts open access traffic data collected by Darmstadt city council. The UI traffic database provides 1-minute interval raw data from inductive-loop detectors and traffic cameras, including vehicle count and road lane occupancy. The database consists of 145 junctions and over 2400 traffic sensors located across Darmstadt city. The data are available from the 27\textsuperscript{th} of June 2017 onwards and refer exclusively to the volume of traffic -- number of cars detected each minute -- allowing no discrimination among different types of vehicles, i.e., cars, buses, lorries, etc. Also, there is no record of accidents or road incidents; however, since the models derived via our proposed techniques reliably capture the normal patterns of traffic (as shown in section \ref{secExpRes}), a substantial discrepancy between sensed and predicted traffic volume is an indication of a traffic incident.

For the purpose of this paper, data were collected between the 27\textsuperscript{th} of June 2017 and the 31\textsuperscript{st} of December 2017 (252,692 readings, each collected 1 minute after the previous one). The data provision outages caused by traffic sensor faults and data service infrastructure maintenance were interpolated using simple cubic interpolation \cite{huynh1993accurate}. More complex data imputation models to address the missing values were not considered at this stage (the interested reader is referred to \cite{qu2009ppca,li2013efficient,sun2006bayesian,xu2014accurate}) because one of the foci of this research is to observe how various prediction models handle real, imperfect data. Traffic data for weekdays as well as weekends were included in the analysis. Raw readings collected every 1 minute were aggregated into 5, 10, 15 and 20 minute intervals, for each traffic detector.

The data used in our analysis, along with detailed maps of the relevant junctions, are fully available on Darmstadt's Urban Institute website, however, they are not downloadable in a straightforward manner. Significant effort was invested to create the necessary scripts for automatically extracting and preprocessing the relevant information from the server. While this is the current practice, \textit{we would like to make a case for streamlining access to such data, especially for the benefit of interested audiences without extensive expertise in data analysis.}

From the point of view of junction topography, the most representative junction from Darmstadt is A13, with five input branches (or lanes) and three output branches -– a structure similar to most of the other junctions in the grid. Specifically, A21 features four input branches and three output branches, whereas A36 receives incoming traffic from three lanes and dispatches outgoing traffic via three lanes. Junction A51 is the most different from A13: while it features four inflow branches and two outflow branches, similarly to A13, there are two additional outflows lanes that are not fitted with sensors -- this provides an additional angle to our comparative analysis, as shown in section \ref{subsecClaim2B}.

\begin{figure*}[h!]
\begin{center}
\begin{subfigure}[t]{0.45\textwidth}
\includegraphics[width=\textwidth]{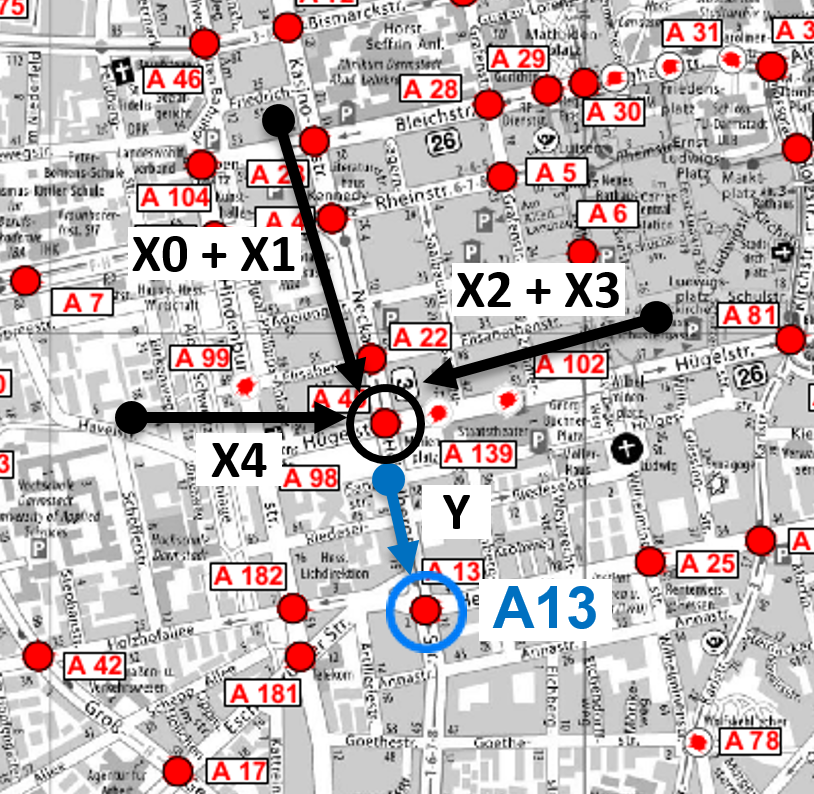}
\caption{Junction A13 map view: X0, ..., X4 are inflow arms and Y is the outflow arm}
\end{subfigure}
\begin{subfigure}[t]{0.46\textwidth}
\includegraphics[width=\textwidth]{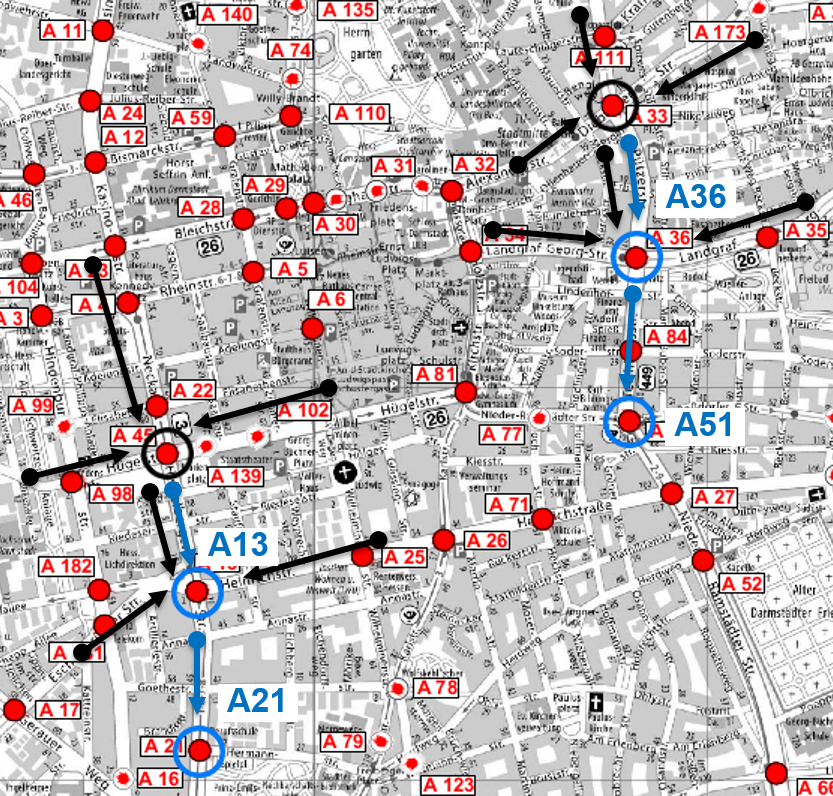}
\caption{Overview of the Darmstadt junctions used for the case study: black arrows indicate inflow of cars into junctions; blue arrows indicate outflow}
\end{subfigure}
\end{center}
\caption{Section of the Darmstadt road network - retrieved from \url{https://darmstadt.ui-traffic.de/faces/TrafficData.xhtml}. Red spots indicate junctions with installed traffic sensors, where traffic is being recorded.}
\label{figA13}
\end{figure*}

\paragraph*{Running example} We will use junction A13 (Fig. \ref{figA13}) located at the center of the Darmstadt road system to illustrate the application of state of the art approaches to traffic modelling, on the one hand, and our proposed method, on the other hand. This will enable us to draw a comparison between the two, on a simple, real world example, thus setting the scene for the more in depth analysis provided in the experimental section.

\section{State of the Art Methods Applied to Traffic Flow Prediction}
\label{secStateOfTheArt}
Given the pronounced seasonality and trend featured by urban traffic flow patterns, specialised approaches, such as triple exponential smoothing (Holt-Winters), are better fit for our research questions than traditional AR(I)MA alternatives. Since our claims are predicated on the inherent advantages of symbolic regression, we also present this method, along with its predecessor, linear regression. The other two categories of methods (probabilistic and non-parametric) are loosely or not at all relevant to this paper's focus and contributions, therefore they are omitted from this section.

\subsection{Holt-Winters}
%[move calculation details to Appendix]

The corner stone of time series forecasting is the assumption that an estimate for the next value in  a time series, $\hat{y}_t$, can be calculated at any moment in time based on the current real value, $y(t)$, and an arbitrary number of past values, $y_{t-1}, ..., y_{t-n}$. Simple moving average would calculate the estimate of the next value as the average of the past $n$ values, while weighted moving average would allocate gradually decreasing weights (in an arithmetic progression) to past values, with the most recent value having the weight $n$ and the value $n$ units of time earlier the weight $1$.
At the same time, exponential smoothing uses exponentially decreasing weights and is widely applied in the case of time series data that presents seasonality. Single exponential smoothing is expressed as:
\begin{equation}
\label{eqSES}
\begin{array}{lcl}
\hat{y}_0  & = & y_0 \\
\hat{y}_t  & = &  \alpha y_t + (1 - \alpha) \hat{y}_{t - 1},\, t>0
\end{array}
\end{equation}
\noindent where $\alpha$ is the data smoothing factor, $0 < \alpha < 1$. The value of $\alpha$ is chosen to give the desired smoothing effect: a value close to one has a smaller smoothing effect and gives greater weight to recent time series values, while a value  close to zero has a larger smoothing effect and is less responsive to recent changes in the time series data.

To facilitate the calculation of estimate $\hat{y}_t$ by numerical means, after performing $n$ recursive substitutions, the estimate in equation \ref{eqSES} becomes:

\begin{equation}
\hat{y}_t = \sum\limits^n_{i = 0}\alpha(1-\alpha)^i y_{t - i}
\end{equation}
\noindent a result called simple exponential smoothing. Indeed, as lag $i$ increases, the weighting of term $y_{t - i}$ decreases, thus extinguishing (or smoothing out) older values in the series.

Single exponential smoothing does not perform well, when there is a trend in the data. In this situation double exponential smoothing is applied. The Holt-Winters method for double exponential smoothing adds the same type of smoothing for the trend $b_t$, defined as the difference between two consecutive values in the time series. Thus, the model becomes:
\begin{equation}
\label{eqDES}
\begin{array}{lcl}
\hat{y}_0  & = & y_0 \\
\hat{y}_t  & = &  \alpha y_t + (1 - \alpha) (\hat{y}_{t - 1} + \hat{b}_{t-1}),\, t>1\\
\hat{b}_t & = & \beta (y_t - y_{t-1}) +(1 - \beta) \hat{b}_{t-1}
\end{array}
\end{equation}
where $\beta$ is the trend smoothing factor.

%\noindent The estimate of the current value in the time series (equation \ref{eqDES}) considering the series' trend (equation \ref{eqTrendComplete}) is called double exponential smoothing. Indeed, as is evident from equation \ref{eqDES}, both the series' values and the series' trend are extinguished as the lag increases.

\begin{figure}[h!]
\begin{center}
\includegraphics[width=.9\textwidth]{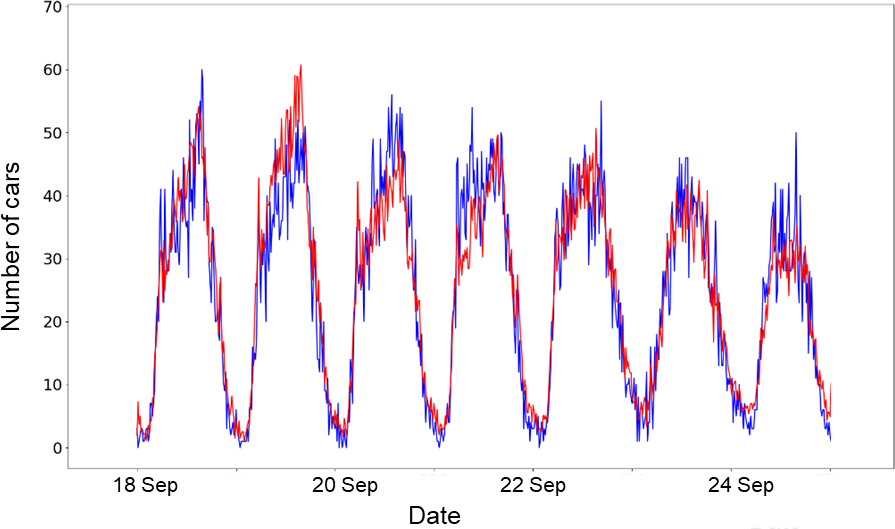}
\end{center}

\caption{A13 junction outflow prediction with Holt-Winters method for 15 minute time windows - level = 0.2, trend = 0.11, seasonality = 0.14; RMSE = 6.56; real - blue, predicted - red}
\label{figHWonA13}
\end{figure}

Finally, seasonality is considered as a repeating pattern in the time series (e.g., Fig. \ref{figHWonA13} shows a one week season on junction A13). The length, $L$, of a season is the number of values in the time series from the beginning to the end of a given pattern.
Triple exponential smoothing with additive seasonality is described by:
\begin{equation}
\label{eqComplete}
\begin{array}{lcl}
\hat{y}_0  & = & y_0 \\
\hat{y}_t  & = &  \alpha ( y_t -c_{t-L}) + (1 - \alpha) (\hat{y}_{t - 1} + \hat{b}_{t-1}),\, t>1\\
\hat{b}_t & = & \beta (y_t - y_{t-1}) +(1 - \beta) \hat{b}_{t-1}\\
c_t & = & \gamma ( y_t - \hat{y}_{t-1} - \hat{b}_{t-1}) + (1- \gamma) c_{t-L}
\end{array}
\end{equation}
\noindent where $\gamma$ is the seasonal smoothing factor. Equations \ref{eqComplete} form the mathematical model for triple exponential smoothing, also known as the Holt-Winters method, where current value, trend and seasonality are smoothed out as the time lag increases. 
\paragraph*{Running example}
Applying the Holt-Winters time series prediction method\footnote{Retrieve Python script from \url{https://grisha.org/blog/2016/02/17/triple-exponential-smoothing-forecasting-part-iii/}} on the A13 junction (Fig. \ref{figA13}) requires the four parameters: $\alpha$ (also called level), $\beta$ (trend) and $\gamma$ (seasonality) and $L$ (season length).   The values used (and appropriately substituted in equations \ref{eqComplete}) are indicated in Table \ref{tabHWSummary}.  The traffic flow characteristics and peak hours of weekdays and weekends are generally different; consequently, one season of traffic flow is represented by one week of data readings.  The obtained prediction for output $y$ is indicated in Fig. \ref{figHWonA13} (red).
We note that the model produced by the Holt-Winters approach only accounts for temporal dependencies, not spatial ones. Additionally, if real output data were missing for a period of time (for example, due to a fault with the sensor), the output would still be predicted but with monotonously decreasing spikes, as recent $y$ terms in the time series gradually  all 
become  zero and older $y$ samples are smoothed out, while inflow terms $x_k$ are not considered at all (as the mathematical model only contains $y$ terms).

%\begin{figure}[h!]
%\includegraphics[width=\textwidth]{A23_HW_5_min.png}
%\caption{A23 junction outflow prediction with Holt-Winters for 5 minute time windows - level = 0.18, trend = 0, seasonality = 0.2; RMSE = 5.7; real - blue, predicted - red}
%\label{figHWonA23}
%\end{figure}

%\begin{figure}[h!]
%\includegraphics[width=\textwidth]{A21_HW_5_min}
%\caption{A21 junction outflow prediction with Holt-Winters for 5 minute time windows - level = 0.8, trend = 0, seasonality = 0.99; RMSE = 3.8; real - blue, predicted - red}
%\label{figHWonA21}
%\end{figure}

\begin{table}[h!]
\caption{Summary of Holt-Winters method experiments on junction A13: 3 weeks training (28/08/2017 - 17/09/2017); 2 weeks testing (18/09/2017 - 01/10/2017)}
\begin{center}

\begin{tabular}{c|c|c|c|c|c}

& Level & Trend & Seasonality & RMSE & MAE, StDev \\
\hline
%A23 1-min& 0.49 & 0.99 & 0 & 1.83 & 1.41 & 1.17& t\\
%A23 5-min& 0.2 & 0.18 & 0 & 5.7 & 4.19 & 3.86 & t\\
%A23 10-min& 0.28 & 0.47 & 0.48 & 9.6 & 6.82 & 6.75 & t\\
%A23 15-min& 0.3 & 0.8 & 0 & 13.61 & 9.45 & 9.79 & t\\
%A23 20-min& 0.07 & 0.88 & 0.79 & 17.20 & 11.84 & 12.49& t\\
%\cline{1-8}
1-min& 0.9   & 0.81 & 0.19 & 1.43 & 1.07, 0.96 \\
5-min& 0.8   & 0.5  & 0.43 & 3.27 & 2.48, 2.14 \\
10-min& 0.43 & 0.78 & 0.41 & 5.04 & 3.81, 3.30 \\
15-min& 0.2  & 0.11 & 0.14 & 6.56 & 5.07, 4.15 \\
20-min& 0.73 & 0.97 & 0.24 & 7.98 & 6.12, 5.11 \\
%\cline{1-8}
%A21 1-min& 0.7 & 0.57 & 0.05 & 1.1 & 0.86 & 0.7 & t\\
%A21 5-min& 0.99 & 0.8 & 0 & 3.8 & 2.97 & 2.37 & t\\
%A21 10-min& 0.71  & 0.27 & 0.62 & 4.2 & 3.3 & 2.61 & t\\
%A21 15-min& 0.81 & 0.39 & 0.93 & 5.58 & 4.38& 3.46 & t\\
%A21 20-min& 0.4 & 0.24 & 0.84 & 6.29 & 4.84 & 4.02 & t\\
\end{tabular}
\end{center}
\label{tabHWSummary}
\end{table}

\subsection{Regression models}
As shown in the previous section, time series modelling estimates output $y(t)$ based on its own past values, $y(t - i)$, where $i$ is an arbitrary lag. In contrast, in regression, the output prediction is constructed as a function, $F$, of $n$ distinct inputs, $x_k$, %The  input set can be extended to include all the past values $x_k(t-i)$ of an arbitrary lag, thus allowing regression to build a model based on up to $n \times L$ input values, where $L$ is the total number of samples collected for both the output and input. 
\begin{equation}
\label{eqRegression}
\hat{y} = F(x_k), k = 1..n
\end{equation}
%\todo{do we  want to keep the lag here in the eq? Probably not.}
Without a built-in selection of the more relevant $x_{k}$, regression models may become overly complex, possibly featuring redundant input.
Considering a model with exogenous input is particularly useful in the sort of traffic analysis where data is collected from real road or in-car sensors, as opposed to being simulated. Specifically, functional regression implicitly accounts for \textit{spatial dependencies} in road traffic, by capturing the way traffic on neighbouring road segments influences the traffic in the location being predicted. Moreover, under realistic conditions, the sensor monitoring the traffic being modelled by $y$ may be faulty or inoperational.
Yet, a prediction can still be obtained by using some form of regression, based on measurements collected from adjacent road segments or junctions (inputs $x_k$ in equation \ref{eqRegression}).

Linear regression assumes that function $F$ is a weighted sum, with $a_{k}$ as weights and $\epsilon$ as the error term: 
\begin{equation}
\label{eqLinRegression}
\hat{y} = \sum _{k=1}^{n}a_{k}x_k + \epsilon,
\end{equation}

\noindent providing a simple, thus computationally straightforward, mathematical model. However, the complexity of urban road configurations often cannot be accurately captured by a linear representation. To address this, symbolic regression removes the restriction of a weighted sum and employs an extended set of operators that can be combined into more complex, nonlinear functions of selected model input $x_k$ (called terminals in this new context). Let us assume a terminal set, $T$, containing all available input at every moment in time alongside real valued numerical constants $\mathfrak{R}$, and an operator set, $O$, comprising arithmetic operators (although, theoretically, there is no limit to the number and type of the elements included):

\begin{equation}
\label{eqSymbRegression}
\begin{array}{rcl}
T & = & \left \{x_k(t), \mathfrak{R} \right \}, k = 1..n\\
O & = & \left \{+, -, *\right \}
\end{array}
\end{equation}

\noindent Running symbolic regression with nothing but the two sets in equation \ref{eqSymbRegression} as a starting point would produce both the structure of function $F$, as well as the regressor coefficients, $a_k$. %\todo{Not really regressor weights in this case anymore, but constants. Also, some constants are allowed in terminal set.} 
Coefficient calculation is a straightforward linear algebra problem, however, \textit{structure selection}, namely deciding on the best combination of operators from set $O$ to connect the regressors within $F$, is more complicated. 

\begin{figure}
\begin{center}
\includegraphics[width=0.6\columnwidth]{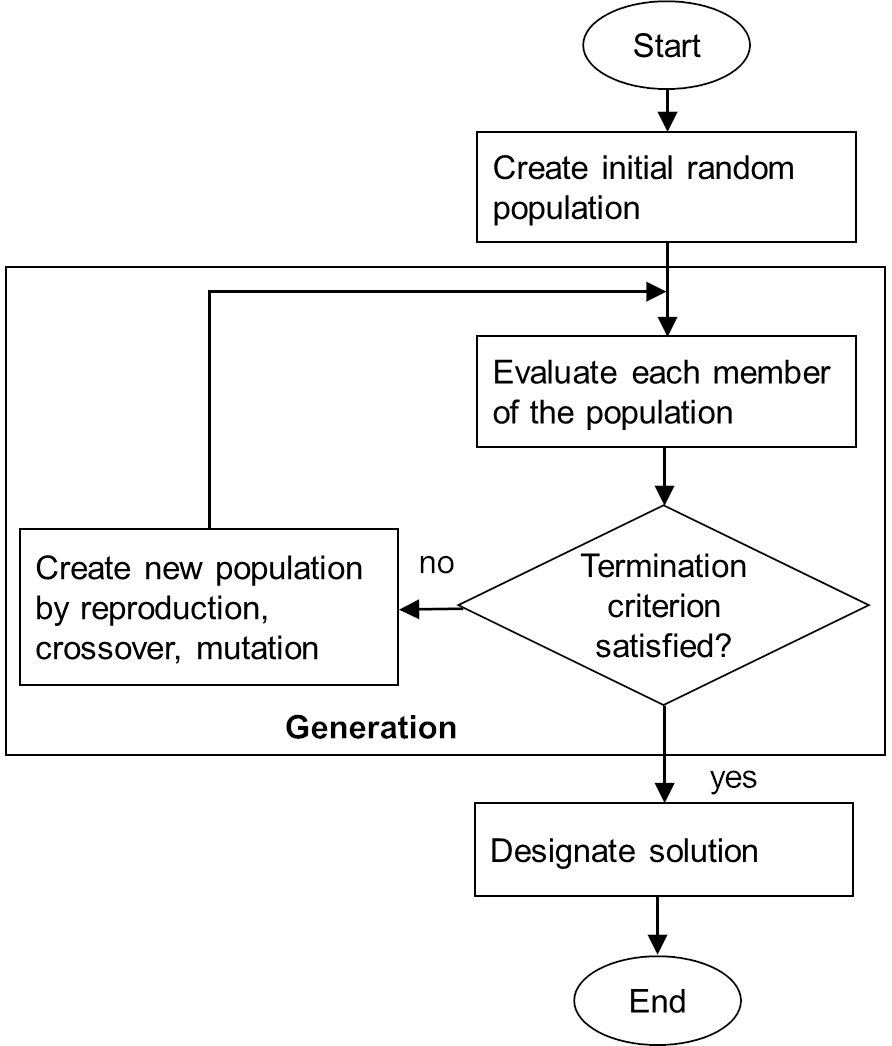}
\caption{A typical Genetic Programming algorithm}
\label{figGP}
\end{center}
\end{figure}

%\subsubsection{Computational symbolic regression}
From an implementation point of view, in symbolic regression, structure selection is achieved by means of Genetic Programming (GP) \cite{koza1992}. In short, GP is an evolutionary technique that starts with a population of randomly generated variants (individuals) of $F$, using the available terminals and operators as building blocks. The individuals can be stored internally as trees, linear sequences or graphs and are improved over generations by subjecting them to genetic operators. Three of the most widely used genetic operators are: \begin{itemize}
\item crossover, requiring two parents and generating two offspring by swapping genetic material (subtrees in the case of tree representation) between the parents
\item mutation, requiring one parent and generating one offspring by applying a small change to the parent (for example change a terminal to another or change an operator to another) 
\item reproduction, requiring one parent and producing one offspring which is an exact copy of the parent.
\end{itemize}
In each generation, parents are selected based on their fitness scores - expressed as numerical measures for tree accuracy (calculated on the available data samples for all featured terminals), but structural simplicity and other criteria may also be of interest. The process of applying genetic operators to selected parents in order to build the new population is repeated, as shown in Fig. \ref{figGP}, until a candidate model for $F$ of satisfactory quality is found or other predefined termination criteria are met.

\begin{figure*}

\begin{subfigure}[t]{0.5\textwidth}
\includegraphics[width=\textwidth]{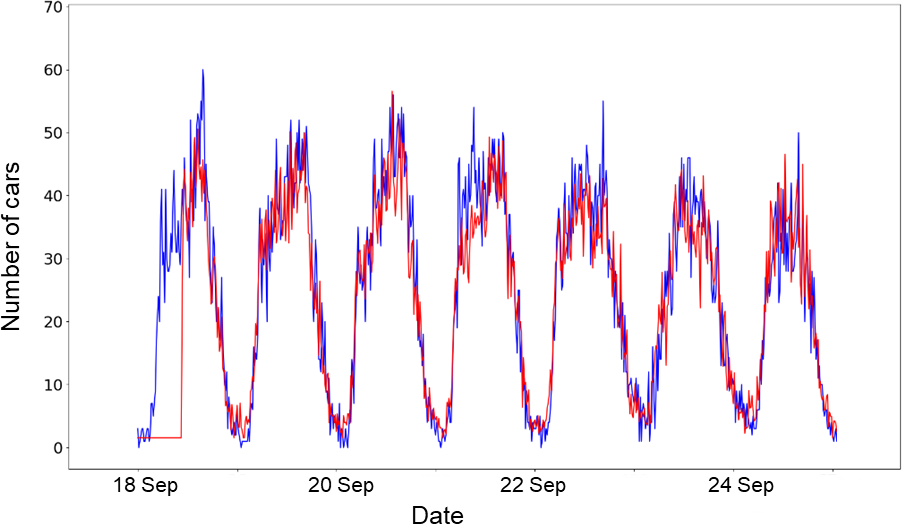}
\caption{Prediction generated with standard linear regression: \\$y(t) = 0.73x_0 + 0.95x_1 + 0.11x_2 + 0.2x_3 + 0.2x_4 + 7.0$ \\RMSE = 7.44}
\end{subfigure}%
\begin{subfigure}[t]{0.51\textwidth}
\includegraphics[width=\textwidth]{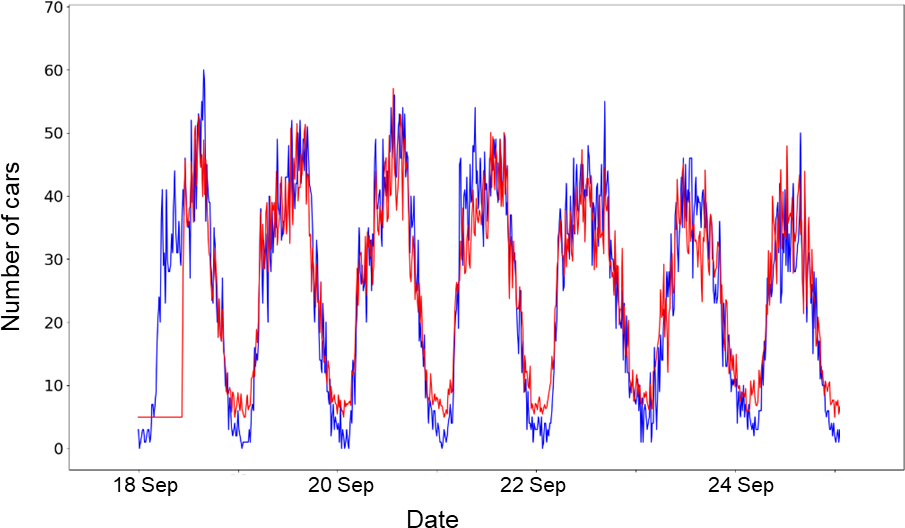}
\caption{Prediction generated with standard symbolic regression: 
%$y(t) = sub(add(add(X0, X2), add(add(add(add(add(add(X1, 0.750), 0.750), 0.750), 0.821), 0.750), 0.750)), mul(add(add(add(add(add(add(X0, X2), X2), mul(X4, -0.622)), X1), mul(X4, -0.622)), X2), 0.147))$
\\$y(t) = 0.853x_0 + 0.853x_1 + 0.559x_2 + 0.183x_4+ 4.571$
\\RMSE = 7.06}
\end{subfigure}%

\caption{A13 junction outflow prediction with numerical regression}
\label{figA13_SymbolicReg_tree_5_minute_no_lag}
\end{figure*}

Besides producing structurally complex models, better capable of capturing realistic road configurations, GP-powered symbolic regression has two other advantages relevant to our case. Firstly, subjecting trees to both crossover and mutation facilitates a balance between exploration (swapping entire sub-structures in order to produce offspring in new areas of the search space) and exploitation (performing small modifications on a fit individual's genome to avoid blockage in local optima). Secondly, given the inherent nature of tree based encoding, the models being evolved become recursive, in that their equivalent mathematical expressions will most likely feature functions of functions. This layered complexity, unavailable in the case of linear regression, will be exploited to our advantage, as shown in section \ref{secSymbRegProposed}.

%Besides our own, other modelling and prediction approaches enlist Genetic Programming as the underlying computational tool, in order to exploit the two advantages mentioned above. 
Genetic programming is being widely used as the underlying computational tool for modelling and prediction, in order to exploit the two advantages mentioned above. 
Specifically, in the medical domain, GP is used for viral protein analysis \cite{fathi2018genetic}, blood glucose dynamics modelling \cite{deFalco2019genetic} and patient data protection \cite{arsalan2017protection}. GP is also deployed to provide numerical insight into industrial systems, such as gas turbines \cite{enriquez2017automatic}, coal processing \cite{garg2017hybrid}, concrete recycling \cite{golafshani2018automatic} and power distribution infrastructure optimisation \cite{pedrino2019islanding} . 
%Other problems successfully tackled with the use of GP are 
GP is successfully being applied in a range of other domains, including
software defect prediction \cite{mauvsa2017co}, image processing \cite{liang2017genetic}, logistics \cite{park2018investigation}, crowd control \cite{hu2018guide} and rainfall prediction \cite{cramer2018decomposition}. These examples are but a few in the vast collection of practical GP implementations, yet they are helpful in conveying the success of this technique for modelling and prediction, both inside and outside the urban transportation context.

\paragraph*{Running example}

Fig. \ref{figA13_SymbolicReg_tree_5_minute_no_lag} shows the predictions obtained with linear regression and traditional symbolic regression (as defined in Equations \ref{eqSymbRegression}).
Both types of numeric regression account for spatial dependencies, but not for temporal ones (the opposite of the Holt-Winters case). This is a direct consequence of their respective mathematical models (equations \ref{eqLinRegression} and \ref{eqSymbRegression}), where exogenous terms (input $x_k$) are factored into the model, measured at the same moment in time as the output (i.e., no historical values). This is of practical value, as demonstrated by the fact that a prediction for output $y$ is being generated even in the time window where the sensor monitoring outflow traffic is down. This is made possible, given that the sensors monitoring the five streams of inflow traffic are still transmitting data, making numeric regression a robust alternative to time series prediction. 
The mathematical expressions for the two regression models are
\begin{equation*}
y(t)  =  0.73 x_0 + 0.95 x_1 + 0.11 x_2 + 0.2 x_3 + 0.2 x_4 + 7.0
\end{equation*}
for linear regression and
\begin{equation*}
y(t) =  0.853 x_0 + 0.853 x_1 + 0.559 x_2 + 0.183 x_4+ 4.571
\end{equation*}
for symbolic regression, respectively. Symbolic regression, after applying arithmetic simplification, also produced a weighted sum of input; however, it eliminated unnecessary input $x_3$ and produced slightly more accurate model than linear regression (RMSE of 7.06 vs 7.44).

%\begin{figure}[h!]
%\includegraphics[width=\textwidth]{A23_LinearRegression_5_min}
%\caption{A23 junction outflow prediction with Linear Regression for 5 minute time windows; real - blue, predicted - red}
%\label{figLinearRegressionA23}
%\end{figure}

%\begin{figure}[h!]
%\includegraphics[width=\textwidth]{A23_SymbolicRegression_5_min.png}
%\caption{A23 junction outflow prediction with Symbolic Regression for 5 minute time windows; real - blue, predicted - red}
%\label{figSymbolicRegressionA23}
%\end{figure}

%\begin{figure}[h!]
%\includegraphics[width=\textwidth]{A21_LinearRegression_5_min}
%\caption{A21 junction outflow prediction with Linear Regression for 5 minute time windows; real - blue, predicted - red}
%\label{figLinearRegressionA21}
%\end{figure}

%\begin{figure}[h!]
%\includegraphics[width=\textwidth]{A21_SymbolicRegression_5_min.png}
%\caption{A21 junction outflow prediction with Symbolic Regression for 5 minute time windows; real - blue, predicted - red}
%\label{figSymbolicRegressionA21}
%\end{figure}

\begin{table*}[h!]
\begin{center}
\caption{Summary of numeric regression experiments for junction A13: 3 weeks training (28/08/2017 - 17/09/2017); 2 weeks testing (18/09/2017 - 01/10/2017) %Includes best RMSE, mean with confidence intervals using normal distribution. p=0.0211 for 10 minutes, p=0.0770 for 15 minutes.
}
\begin{tabular}{c|c|c|c|c|c|c|}
& \multicolumn{2}{c|}{LR} & \multicolumn{2}{c|}{SR} & \multicolumn{2}{c|}{SL} \\
min&RMSE & MAE, & RMSE & MAE, StDev & RMSE & MAE, StDev \\
&&StDev&RMSE $\pm$ CI & MAE $\pm$ CI&RMSE $\pm$ CI & MAE $\pm$ CI\\
\hline
%A23 1-min& 1.76 & 1.37 & 1.11 & 1.74 & 1.34 & 1.1 \\
%A23 5-min& 5.93 & 4.64 & 3.69 & 5.69 & 4.37 & 3.64 \\
%A23 10-min& 10.42 & 8.12 & 6.53 & 9.82 & 7.44 & 6.41 \\
%A23 15-min& 14.85 & 11.45 & 9.46 & 13.95 & 10.71 & 8.94 \\
%A23 20-min& tick & cross& tick& cross& tick& cross\\
%\cline{1-7}
1& 1.47 & 1.14, 0.94 & 1.48 & 1.15, 0.93 & 1.41 & 1.09, 0.89 \\
 &  & & $1.55 \pm 0.01$ & $1.18 \pm 0.005$ & $1.52 \pm 0.01 $ & $1.16 \pm 0.008$ \\
5& 3.58 & 2.8, 2.24 & 3.53 & 2.72, 2.25 & 3.4 & 2.66, 2.13 \\
 &  &  & $3.65 \pm 0.03$ & $2.83 \pm 0.03$ & $3.6 \pm 0.03$ & $2.8 \pm 0.03$ \\
% &  &  &  &  &  &  & 2.78, 4.85 &  \\
10& 5.62 & 4.37, 3.54 & 5.44 & 3.91, 3.78 & 5.29 & 3.97, 3.49 \\
 &  & & $5.61 \pm 0.04$ & $4.24 \pm 0.06$ & $5.53 \pm 0.03$ & $4.18 \pm 0.06$ \\
15& 7.44 & 5.85, 4.6 & 7.06 & 5.21, 4.8 & 6.91 & 4.68, 5.09 \\
& &  &  $7.33 \pm 0.04$ & $5.48 \pm 0.08$ &  $7.28 \pm 0.05$ & $5.22 \pm 0.09$ \\
% &  &  &  &  &  &  & 5.62, 9.80 &  &  \\
 
%20& tick & cross& tick& x & x & x & cross& tick& cross\\
%\cline{1-7}
%A21 1-min& 1.12 & 0.89 & 0.67 & 1.08 & 0.84& 0.68 \\
%A21 5-min& 2.8 & 2.17 & 1.76 & 2.6 & 1.97 & 1.69 \\
%A21 10-min& 4.39 & 3.32 & 2.86 & 3.92 & 2.9 & 2.64 \\
%A21 15-min& 5.8 & 4.37 & 3.82 & 5.21 & 3.78 & 3.58 \\
%A21 20-min& tick & cross& tick& cross& tick& cross\\
\end{tabular}
\end{center}
\label{tabLinearRegression}
\end{table*}

\section{Specialised Symbolic Regression for Traffic Flow Prediction}
\label{secSymbRegProposed}
\begin{figure}[hbt]
\center
\includegraphics[width=0.9\textwidth]{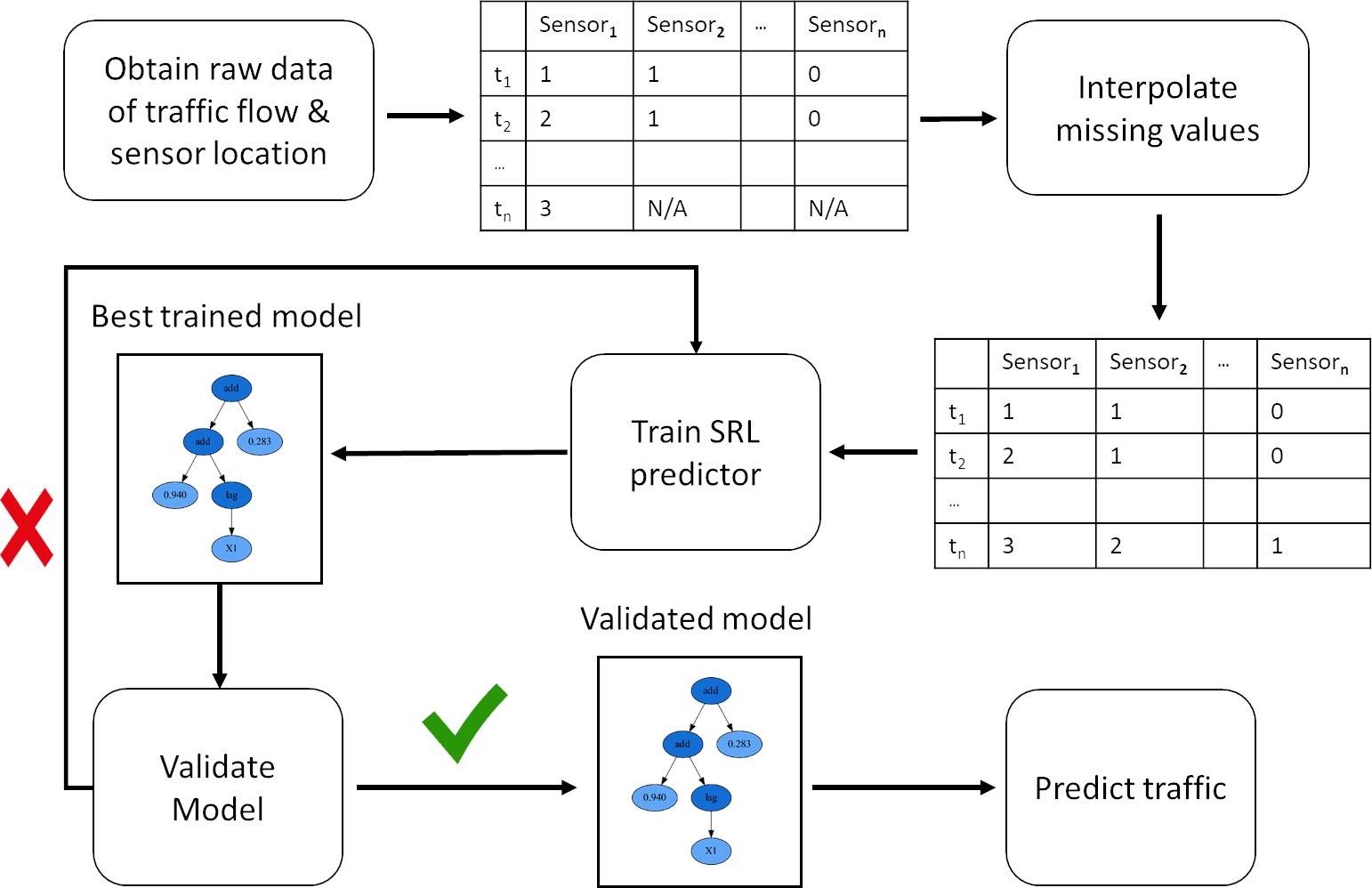}
\caption{Work process for Symbolic Regression with Lag algorithm}
\label{figwp}
\end{figure}

 The proposed work process is illustrated in Figure~\ref{figwp}. After obtaining the raw data and interpolating missing values, the training and validation of models follows. The cycle is repeated until a model of desired accuracy is obtained, which is subsequently used for prediction.
 
\noindent Traffic modelling on the Darmstadt case study was performed using the \verb|gp learn| API\footnote{\url{https://gplearn.readthedocs.io/en/stable/intro.html}}, available in Python. The authors wrote a script, using the previously mentioned API, consisting of three sections: establishing the terminal and operator sets to be used by the symbolic regression based estimator, configuring all relevant parameters of the estimator and evaluating the resulting model.
\paragraph*{Terminal and operator sets} To account for \textit{temporal dependencies} in road traffic, it is necessary to consider lagged inputs, namely values collected in the past from road segments neighbouring the one being predicted. This could be achieved by enriching terminal set $T$ accordingly:
\begin{equation}
\label{eqSymbRegressionWithLag}
\begin{array}{ccl}
T & = & \left \{x_k(t), x_k(t - 1), ..., x_k(t-L), \mathfrak{R}\right \}, k = 1..n\\
O & = & \left \{+, -, *\right \}
\end{array}
\end{equation}
However, the optimal value for $L$, namely the size of the time window where past input samples influence the output significantly, is not known \textit{a priori}. A poorly chosen maximum lag may lead to either insufficient or exaggerated number of terminals. In the former case, the model will be inaccurate. In the latter case, the insignificant terminals may be computationally expensive to eliminate from the final model. To prevent both scenarios, we have decided against the inclusion of delayed terminals and, instead, we have introduced a lag function in the operator set (also attempted in \cite{mcconaghy2000functional}, on a simplified, polynomial scenario): 
\begin{equation}
\label{eqSymbRegTerminalSetMinAndOpWithLag}
\begin{array}{ccl}
T & = & \left \{x_k(t), \mathfrak{R} \right \}, k = 1..n \\
O & = & \left \{+, -, *, lagOne\right \}, lagOne(x_k(t)) = x_k(t - 1)
\end{array}
\end{equation}
Another advantage of this approach is that $lagOne$ is the only addition needed to operator set $O$. This is because symbolic regression includes operators recursively in the models it evolves, therefore creating implicit scope for utilising lags of higher orders (from 1 to the maximum allowed depth of regression trees),
$lagOne^d(x_k(t)) = x_k(t-d)$.

\paragraph*{Estimator configuration} The key parameters\footnote{A full description of the algorithm's configuration is available at \url{https://gplearn.readthedocs.io/en/stable/reference.html} } used to configure the \verb|gp lab| symbolic regression based estimator are the population size, the number of generations, the probabilities of applying crossover and mutation at each generation, the operator set (given in equation \ref{eqSymbRegTerminalSetMinAndOpWithLag}) and the fitness metric. As a bloat control mechanism, the symbolic regression algorithm uses a parsimony coefficient, a dedicated parameter that penalises oversized trees by reducing their fitness, as opposed to eliminating them from the population entirely. The penalty is calculated with respect to the covariance between the tree's fitness and its size. The values used for all algorithm parameters are presented and analysed in the experimental section \ref{secExpRes}. 

\paragraph*{Model evaluation} Besides the classic Root Mean Square Error (RMSE), our Python script assesses model quality by using the coefficient of determination, $R^2$. This is calculated as shown in equation \ref{eqCoefOfDeterm}, where $N$ is the number of output data samples.
\begin{equation}
\label{eqCoefOfDeterm}
\begin{split}
R^2 = 1 - \frac{\sum\limits_{t}(\hat{y}(t) - y(t))^2}{\sum\limits_{t}(y(t) - M)^2}\\
M = \frac{\sum\limits_{t} y(t)}{N}
\end{split}
\end{equation}
A perfect estimation would yield a value of 1 for $R^2$, however, a model may be considered of good quality if it generates a value close to 1, namely, if the approximation error is very small in comparison to the output's variance. This level of subtlety in evaluating the quality of the prediction is not available by calculating RMSE alone.

\begin{figure}[h!]
\begin{center}
\begin{subfigure}[t]{0.5\textwidth}
\includegraphics[width=1\linewidth]{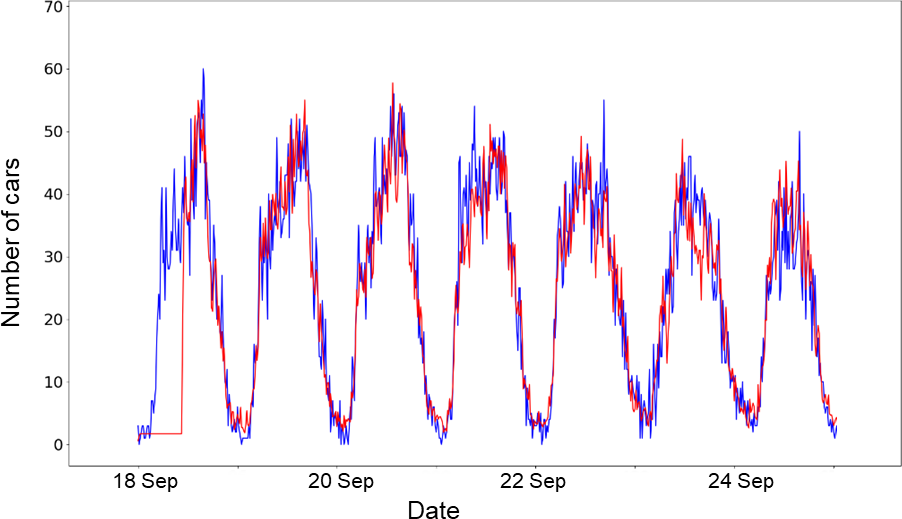}
\caption{Prediction generated with symbolic regression with lag:
%\\$y(t) = add(add(0.444, lag(add(0.444, mul(0.420, add(add(0.444, add(0.351, X0)), lag(add(0.444, mul(0.420, add(add(0.444, mul(0.420, lag(add(X2, X0)))), add(0.351, X0)))))))))), add(0.152, add(X1, mul(0.420, add(X0, X2)))))$ 
\\$y(t) = 0.42x_0 + 0.42 lag(x_0) + 0.176 lag^2(x_0) + 0.074 lag^3(x_0) + x_1 + 0.42x_2 + 0.074 lag^3(x_2) + 0.596$
\\RMSE = 6.91}
\end{subfigure}
\begin{subfigure}[t]{0.3\textwidth}
\includegraphics[width=1\linewidth]{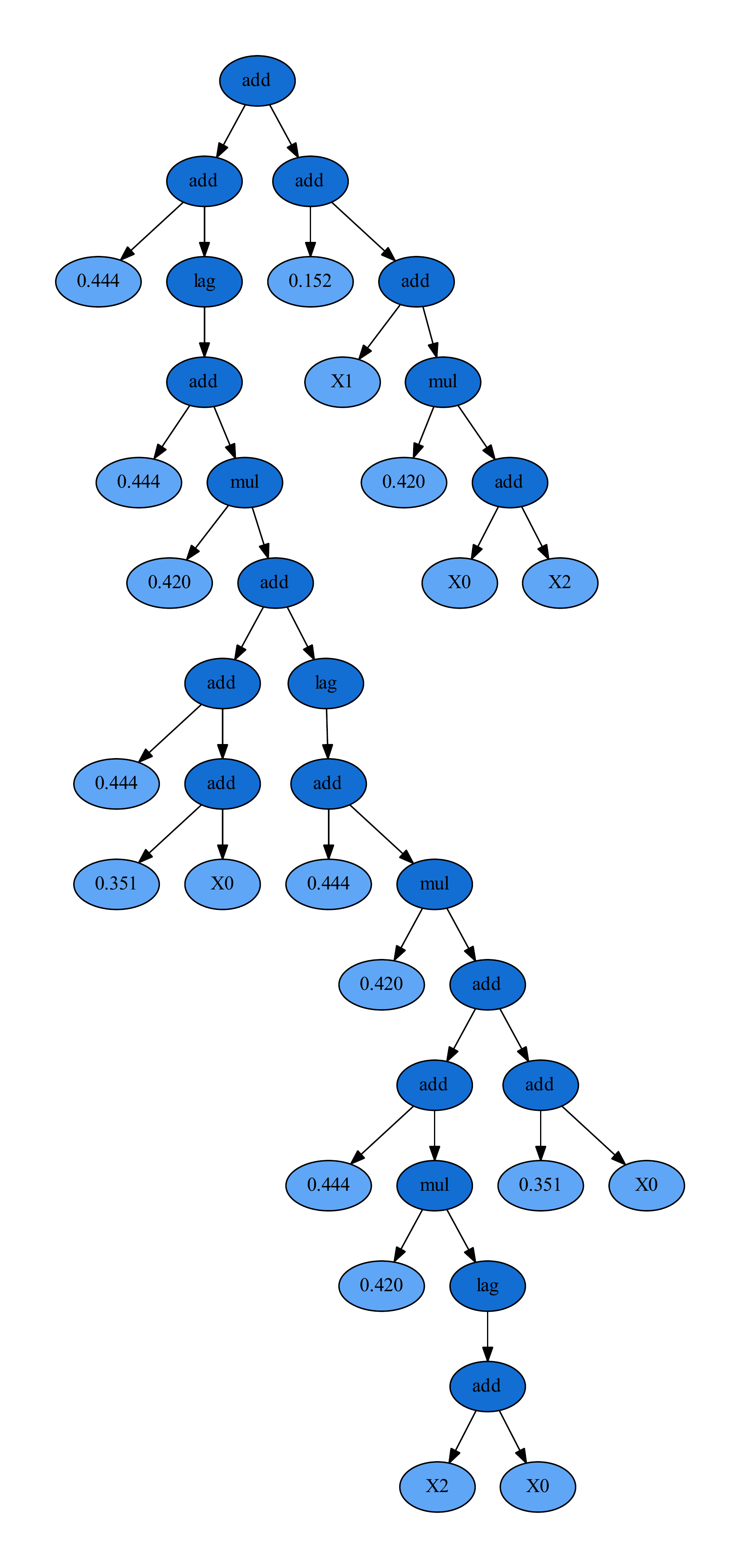}
\caption{Prediction tree generated by symbolic regression with lag}
\end{subfigure}
\end{center}
\caption{A13 junction outflow prediction by symbolic regression with lag}
\label{figA13_SymbRegressionWithLag}
\end{figure}

\paragraph*{Running example} 
Fig. \ref{figA13_SymbRegressionWithLag} illustrates the prediction for junction A13's outflow arm, $y$, produced by means of symbolic regression with a lag operator.
Our proposed approach accounts for spatial dependencies as well as temporal ones, an improvement over the three alternatives presented in the previous sections. In specific terms, the symbolic regression variant we propose (equation \ref{eqSymbRegTerminalSetMinAndOpWithLag}) considers exogenous terms (input $x_k$) over a variable-sized window of past samples (given the inclusion of the lag operator in set $O$).  Another notable advantage that symbolic regression with a lag operator holds over linear regression is the unsupervised exclusion of irrelevant input from the final model. The model  
$y(t) = 0.42 x_0(t) + 0.42 x_0(t-1) + 0.176 x_0(t-2) + 0.074 x_0(t-3) + x_1 + 0.42 x_2 + 0.074 x_2(t-3) + 0.596$ of Fig. \ref{figA13_SymbRegressionWithLag}, when compared to the model derived using simple symbolic regression (see Fig. \ref{figA13_SymbolicReg_tree_5_minute_no_lag}(b)), is more sophisticated and at the same time more parsimonious, as it makes use of up to three historical values for $x_0, x_1, x_2$ and no values of $x_4$, in addition to $x_3$ already eliminated by simple symbolic regression.  

%\begin{table*}[ht]
%\begin{tabular}{c|c|c|c|c|c|c|c|c|c|c|}
%& \multicolumn{2}{c|}{A13} & \multicolumn{2}{c|}{A21}& \multicolumn{2}{c|}{A36} & \multicolumn{2}{c|}{A51} & \multicolumn{2}{c|}{Total} \\
%Count & 5min & 15min & 5min & 15min & 5min & 15min & 5min & 15min & 5min & 15min \\
%0 & 2 & 2 & 2 & 0 & 38 & 37 & 39 & 43 & 81 & 82 \\
%1 & 39 & 19 & 33 & 3 & 8 & 7 & 4 & 1 & 84 & 30 \\
%2 & 5 & 8 & 6 & 33 & 3 & 4 & 5 & 2 & 19 & 47 \\
%3 & 1 & 9 & 2 & 4 & 0 & 0 & 1 & 0 & 4 & 13 \\
%4 & 0 & 4 & 1 & 1 & 0 & 1 & 1 & 1 & 2 & 7 \\
%5 & 3 & 1 & 0 & 0 & 1 & 0 & 0 & 1 & 4 & 2 \\
%6 & 0 & 3 & 3 & 1 & 0 & 0 & 0 & 1 & 3 & 5 \\
%7 & 0 & 2 & 0 & 1 & 0 & 0 & 0 & 1 & 0 & 4 \\
%8 & 0 & 1 & 0 & 1 & 0 & 0 & 0 & 0 & 0 & 2 \\
%9 & 0 & 0 & 0 & 2 & 0 & 0 & 0 & 0 & 0 & 2 \\
%10 & 0 & 0 & 0 & 1 & 0 & 1 & 0 & 0 & 0 & 2 \\
%11 & 0 & 0 & 0 & 0 & 0 & 0 & 0 & 0 & 0 & 0 \\
%12 & 0 & 0 & 2 & 0 & 0 & 0 & 0 & 0 & 2 & 0 \\
%13 & 0 & 0 & 0 & 1 & 0 & 0 & 0 & 0 & 0 & 1 \\
%14 & 0 & 1 & 1 & 0 & 0 & 0 & 0 & 0 & 1 & 1 \\
%15 & 0 & 0 & 0 & 2 & 0 & 0 & 0 & 0 & 0 & 2 \\
%\end{tabular}
%\caption{Count of Lag operators in the best 50 models for junctions A13, A21, A36, A51 for 5 and 15 minute intervals.}
%\label{tabLagCount}
%\end{table*}

\section{Experimental Results}
\label{secExpRes}
% \begin{table}[h!]
% \begin{tabular}{c|c|c|c|c|c|c}
% Claim &1 & 2a & 2b & 2c & 3a & 3b\\
% HW& tick & cross& tick& cross& tick& cross\\
% LR & tick & cross& tick& cross& tick& cross\\
% SR & tick & cross& tick& cross& tick& cross\\
% SL & tick & cross& tick& cross& tick& cross\\
% \end{tabular}
% \caption{Summary of experiments [placeholder - tick/cross to be replaced with actual symbols according to experimental results + more rows to be added for different parameter configurations for each method]}
% \label{tabExpSummary}
% \end{table}
Our
results support the three main claims made in the introduction, highlighting the superiority of our traffic prediction strategy based on symbolic regression featuring a lag operator (SL) over the other three more traditional approaches considered in this paper, namely Holt-Winters (HW), linear regression (LR) and standard symbolic regression (SR). First, the detailed experiments performed on junction A13 are analysed, building on the running example sections above. We then apply the insight gained from the junction A13 case study to the system of junctions highlighted in Fig. \ref{figA13}(b) and analyse the effects. 

On a methodology related note, we only provide p-values for our claims where these are of statistical relevance. Another methodology related decision  -- i.e., to calculate model errors as whole numbers, rather than percentages -- is also necessary, as there are many recordings of zero volume (i.e periods of time with no traffic) in the data set.

\subsection {Claim 1: Our approach features high tolerance to sensor faults}

\begin{figure}
\begin{center}
\includegraphics[width=0.42\textwidth]{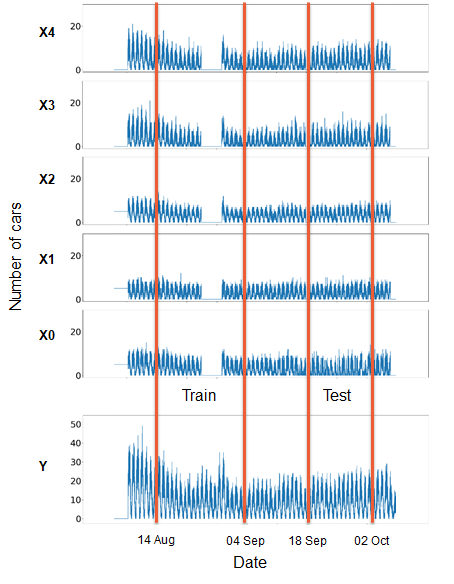}
\end{center}

\caption{Traffic flow for all input and output for junction A13: red lines indicate data used for some of the experiments run in support of claim 1: 3 weeks training (14/08/2017 0:00 - 03/09/2017 23:59); 2 weeks testing (18/09/2017 0:00 - 01/10/2017 23:59)}
\label{figA13MissingData}
\end{figure}

%To test whether the selected methods can produce models with high tolerance to severe sensor faults, we trained the models on 3 weeks of data that included a large proportion of missing sensor readings (4.5 consecutive days of data missing) (see Figure \ref{figA13MissingData}). The models were then tested on 2 weeks of data to evaluate accuracy. The results are shown in tables \ref{tabHWSummaryMissingData}, \ref{tabLinearRegressionMissingData} and \ref{tabSymbolicRegressionMissingData}.
Holt-Winters, linear regression and symbolic regression with lag were all considered in the investigation of claim 1. The three algorithms were run on a continuous 3 week training window ($28^{\text{th}}$ of August to $17^{\text{th}}$ of September in Fig. \ref{figA13MissingData}) and, subsequently, on a gapped 3 week training window ($14^{\text{th}}$ of August to $3^{\text{rd}}$ of September in Fig. \ref{figA13MissingData}, with samples between $24^{\text{th}}$ of August to $28^{\text{th}}$ of August missing) \footnote{Our claim 1 is made considering a substantial percentage (over 25\%) of missing data, regardless of whether that missing data are grouped together or scattered across several gaps throughout the training interval. We demonstrate the feasibility of our approach on an interval with one gap,  but the experiment could be run similarly on an interval with several gaps.}. All models were tested on the same 2 week window (samples between $18^{\text{th}}$ of September to $1^{\text{st}}$ of October in Fig. \ref{figA13MissingData}). The sampling intervals on the training and testing data sets were 1, 5, 10 and 15 minutes. All experiments relevant to Claim 1 were run on junction A13. 

The performance of the models produced by all three methods, trained on a continuous window as well as on a gapped window, is presented in tables \ref{tabHWSummaryMissingData}, \ref{tabLinearRegressionMissingData} and \ref{tabSymbolicRegressionMissingData}, respectively. The tables report the Root Mean Squared Error (RMSE) and Mean Average Error (MAE), accompanied by either the Standard Deviation (StDev) or the Confidence Interval (CI), calculated on the testing data set. In the case of symbolic regression with lag (Table \ref{tabSymbolicRegressionMissingData}), rows titled ``best'' refer to the most accurate tree obtained after 50 runs of the SL algorithm, for which we report the MAE and the StDev. Rows titled ``all'' refer to the average performance over the set of 50 most accurate trees (each obtained in a separate run), reported by means of the MAE with an indication of the confidence interval.

\paragraph*{Discussion} Results indicate that Holt-Winters provides the most accurate models, mainly due to the fact that this time-series based approach is insensitive to missing data from input lanes, as it is not using this input. Overall symbolic regression with lag performs better than linear regression and yields models that are comparable to that produced by Holt-Winters (the most significant difference, in terms of RMSE, is recorded for the 15 minute sampling interval, on gapped training data, where SL errs more than HW by 1.36 cars). When comparing the performance of SL on continuous vs gapped training data, statistical testing rejected the null hypothesis: p values are under 0.05, indicating that the two sets of RMSE scores are significantly different. However, the variations are small, with the highest loss in accuracy, caused by training on gapped data, being 1.86 cars on a 15 minute sampling interval, in the ``all'' measurements of the RMSE value. This validates our claim 1 in the sense that, when trained on gapped data as opposed to continuous data, the loss in accuracy of the model produced by symbolic regression with lag is under 2 cars over 15 minutes.
\begin{table}[h!]
\begin{center}
\caption{Holt-Winters method on junction A13 with continuous (left) and gapped training windows (right). There is an overlap with Table \ref{tabHWSummary} to improve readability}
\begin{tabular}{c|c|c|c|c|}
& \multicolumn{2}{c|}{Continuous} & \multicolumn{2}{c|}{Gapped}  \\
& RMSE & MAE, StDev & RMSE & MAE, StDev \\
\hline
1 min& 1.43 & 1.07, 0.96 & 1.48 & 1.09, 0.99 \\
5 min& 3.27 & 2.48, 2.14 & 3.47 & 2.58, 2.31 \\
10 min& 5.04 & 3.81, 3.30 & 5.3 & 3.95, 3.54 \\
15 min& 6.56 & 5.07, 4.15 & 7.16 & 5.33, 4.78 \\
\end{tabular}
\label{tabHWSummaryMissingData}
\end{center}
\end{table}

\begin{table}[h!]
\begin{center}
\caption{Linear regression on junction A13 with continuous (left) and gapped training windows (right)}
\begin{tabular}{c|c|c|c|c|}
& \multicolumn{2}{c|}{Continuous} & \multicolumn{2}{c|}{Gapped} \\
&RMSE & MAE, StDev & RMSE & MAE, StDev \\
\hline
1 min& 1.47 & 1.14, 0.94 & 1.48 & 1.2, 0.88 \\
5 min& 3.58 & 2.8, 2.24 & 4.02 & 3.4, 2.15 \\
10 min& 5.62 & 4.37, 3.54 & 6.84 & 5.8, 3.64 \\
15 min& 7.44 & 5.85, 4.6 & 9.54 & 8.09, 5.04 \\
\end{tabular}
\label{tabLinearRegressionMissingData}
\end{center}
\end{table}

\begin{table*}[h!]
\begin{center}
\caption{Symbolic regression with lag on junction A13 with continuous (left) and gapped training windows (right); \\p = 0.3$\times 10^{-10}$ for 10 and 15 min sampling intervals}%p=0.00000000003019 for 10 minutes, p=0.00000000003018 for 15 minutes - for both cases the populations are distinct.}
\begin{tabular}{cc|c|c|c|c|}
& &\multicolumn{2}{c|}{Continuous} & \multicolumn{2}{c|}{Gapped} \\
&&RMSE & MAE, StDev & RMSE & MAE, StDev \\
&&RMSE $\pm$ CI & MAE $\pm$ CI&RMSE $\pm$ CI & MAE $\pm$ CI\\
\hline
1 min & best& 1.41 & 1.09, 0.89 & 1.44 & 1.17, 0.83 \\
 &all & $1.52 \pm 0.01 $ & $1.16 \pm 0.008$  & $1.55 \pm 0.009$ & $1.27 \pm 0.009$   \\
5 min & best& 3.4 & 2.66, 2.13 & 3.96 & 3.29, 2.21 \\
 & all & $3.6 \pm 0.03$ & $2.8 \pm 0.03$   & $4.04 \pm 0.03$ & $3.38 \pm 0.03$ \\
10 min & best & 5.29 & 3.97, 3.49 & 6.3 & 5.27, 3.45 \\
 & all & $5.53 \pm 0.03$ & $4.18 \pm 0.06$  & $6.61 \pm 0.06$ & $5.59 \pm 0.06$  \\
15 min & best & 6.91 & 4.68, 5.09 & 8.52 & 7.1, 4.7 \\
 & all & $7.28 \pm 0.05$ & $5.22 \pm 0.09$  & $9.14 \pm 0.11$ & $7.72 \pm 0.096$ \\
\end{tabular}
\label{tabSymbolicRegressionMissingData}
\end{center}
\end{table*}

\iffalse
############################################################
\begin{figure}[h!]
\includegraphics[width=\textwidth]{A13_SymbolicRegression_5_min_missing.png}
\caption{A13 junction outflow prediction with Symbolic Regression for 5 minute time windows with 6 days of data missing from training set; real - blue, predicted - red}
\label{figLinearRegressionA13Missing}
\end{figure}

\begin{figure}[h!]
\includegraphics[width=\textwidth]{A13_SymbolicRegression_10_min_missing.png}
\caption{A13 junction outflow prediction with Symbolic Regression for 10 minute time windows with 6 days of data missing from training set; real - blue, predicted - red}
\label{figLinearRegressionA13Missing}
\end{figure}

\begin{figure}[h!]
\includegraphics[width=\textwidth]{A13_SymbolicRegression_15_min_missing.png}
\caption{A13 junction outflow prediction with Symbolic Regression for 15 minute time windows with 6 days of data missing from training set; real - blue, predicted - red}
\label{figLinearRegressionA13Missing}
\end{figure}
##############################################################
\fi
\subsection{Claim 2a: Our models are robust and reliable, irrespective of training window}
%[vary the size of the training window - small, medium, large - table with RMSE and MAE for Holt-Winters, linear regression, symbolic regression with lag ]
Linear and symbolic regression with lag were run on continuous training windows of 1 week ($11^{\text{th}}$ of September to $17^{\text{th}}$ of September in Fig. \ref{figA13MissingData}), 2 weeks ($4^{\text{th}}$ of September to $17^{\text{th}}$ of September) and 3 weeks ($28^{\text{th}}$ of August to $17^{\text{th}}$ of September). The resulting models were tested on the same continuous training data set as in the previous section ($18^{\text{th}}$ of September to $1^{\text{st}}$ of October). The training and testing data sets were sampled every 1, 5, 10 and 15 minutes. The model produced by Holt-Winters was trained only on 2 or 3 weeks, respectively, with the same testing window as in the case of the other two methods. The tests were executed on junction A13. The metrics reported in tables \ref{tabHWVariableTrainingWindow}, \ref{tabLinearRegressionVariableTrainingWindow} and \ref{tabSymbolicRegressionVariableTrainingWindow} have the same significance as previously.
\paragraph*{Discussion} Holt-Winters can not be trained on a single week of data, as that is the length of one season. At least two seasons are necessary to provide sufficient data for an effective use of Holt-Winters. Linear regression provides a suitable model even on the shortest training window considered, with no significant change in accuracy as the number of training samples is increased to 2 and 3 weeks. Symbolic regression with lag fares marginally better than linear regression on 1 week of training and yields results comparable to Holt-Winters on the other two training windows - the highest accuracy deficit is recorded on 3 weeks of training, with data sampled every 15 minutes, when SL produces a RMSE value that is 0.35 cars higher than that of Holt-Winters. When comparing the performance of SL on the three training window lengths, statistical testing supports the null hypothesis: p values are substantially higher than 0.05, indicating that the three sets of RMSE scores are not significantly different. Our claim 2a is thus validated, in that symbolic regression with lag produces models after one week of training with similar accuracy to the ones trained on two or three weeks worth of data, provided that the training set is not missing a substantial amount of samples. 
\begin{table}[h!]
\begin{center}
\caption{Holt-Winters method on junction A13 with varying training windows. The overlap with Table \ref{tabHWSummary} is for the purpose of improving readability}
\begin{tabular}{c|c|c|c|c|}
& \multicolumn{2}{c|}{2 weeks} & \multicolumn{2}{c|}{3 weeks} \\
& RMSE & MAE, StDev & RMSE & MAE, StDev \\
\hline
1 min& 1.5 & 1.11, 1.00 & 1.43 & 1.07, 0.96 \\
5 min& 3.39 & 2.57, 2.22 & 3.27 & 2.48, 2.14 \\
10 min& 5.05 & 3.82, 3.3 & 5.04 & 3.81, 3.30 \\
15 min& 6.47 & 5.00, 4.11 & 6.56 & 5.07, 4.15 \\
\end{tabular}
\label{tabHWVariableTrainingWindow}
\end{center}
\end{table}

\begin{table*}[h!]
\begin{center}
\caption{Linear regression on junction A13 with varying training windows}
\begin{tabular}{c|c|c|c|c|c|c|}
& \multicolumn{2}{c|}{1 week} & \multicolumn{2}{c|}{2 weeks}  & \multicolumn{2}{c|}{3 weeks} \\
&RMSE & MAE, StDev & RMSE & MAE, StDev & RMSE & MAE, StDev \\
\hline
1 min& 1.47 & 1.13, 0.94 & 1.48 & 1.12, 0.96 & 1.47 & 1.14, 0.94 \\
5 min& 3.49 & 2.6, 2.33 & 3.54 & 2.61, 2.4 & 3.58 & 2.8, 2.24 \\
10 min& 5.41 & 3.87, 3.78 & 5.48 & 3.87, 3.88 & 5.62 & 4.37, 3.54 \\
15 min& 7.07 & 4.91, 5.08 & 7.15 & 4.94, 5.18 & 7.44 & 5.85, 4.6 \\
\end{tabular}
\label{tabLinearRegressionVariableTrainingWindow}
\end{center}
\end{table*}

\begin{table*}[h!]
\begin{center}
\caption{Symbolic regression with lag on junction A13 with varying training windows; 
p=0.3329 (1 week vs 2 weeks, 10 min); p = 0.1413 (1 week vs 3 weeks, 10 min); p = 0.6627 (2 weeks vs 3 weeks, 10 min); p = 0.3007 (1 week vs 2 weeks, 15 min); p = 0.2226 (1 week vs 3 weeks, 15 min); p = 0.5201 (2 weeks vs 3 weeks, 15 min); p values for 1 min and 5 min sampling intervals are not included as they are less statistically conclusive}
\begin{tabular}{cc|c|c|c|c|}
&& \multicolumn{2}{c|}{1 week } & \multicolumn{2}{c}{2 weeks}  \\
&&RMSE & MAE, StDev & RMSE & MAE, StDev  \\
&&RMSE $\pm$ CI & MAE $\pm$ CI&RMSE $\pm$ CI & MAE $\pm$ CI \\
\hline
1 min& best & 1.39 & 1.07, 0.88 & 1.38 & 1.07, 0.88  \\
 &all & $1.49 \pm 0.02$ & $1.14 \pm 0.02$  & $1.51 \pm 0.02$ & $1.14 \pm 0.01$ \\
5 min& best & 3.3 & 2.38, 2.29 & 3.28 & 2.38, 2.26  \\
 & all &$3.54 \pm 0.04$ & $2.59 \pm 0.03$  & $3.51 \pm 0.04$ & $2.57 \pm 0.03$    \\
10 min&best & 5.23 & 3.59, 3.81 & 5.09 & 3.44, 3.75  \\
 &all& $5.56 \pm 0.06$ & $3.92 \pm 0.05$ & $5.52 \pm 0.07$ & $3.86 \pm 0.06$   \\
15 min& best &6.92 & 4.62, 5.15 & 6.9 & 4.55, 5.19   \\
 &all & $7.35 \pm 0.06$ & $4.99 \pm 0.05$ & $7.31 \pm 0.06$ & $4.97 \pm 0.06$   \\
\end{tabular}

\vspace{10pt}

\begin{tabular}{cc|c|c|}
&&  \multicolumn{2}{c}{3 weeks} \\
&&  RMSE & MAE, StDev \\
&&RMSE $\pm$ CI & MAE $\pm$ CI\\
\hline
1 min& best &   1.41 & 1.09, 0.89 \\
 &all &   $1.52 \pm 0.01$ & $1.16 \pm 0.008$ \\
5 min& best &  3.4 & 2.66, 2.13 \\
 & all & $3.6 \pm 0.03$ & $2.8 \pm 0.03$ \\
10 min&best &  5.29 & 3.97, 3.49 \\
 &all&  $5.53 \pm 0.03$ & $4.18 \pm 0.06$ \\
15 min& best &  6.91 & 4.68, 5.09 \\
 &all &   $7.28 \pm 0.05$ & $5.22 \pm 0.09$ \\
\end{tabular}
\label{tabSymbolicRegressionVariableTrainingWindow}
\end{center}
\end{table*}

\subsection{Claim 2b: Our models are flexible enough for urban traffic}
\label{subsecClaim2B}
To investigate claim 2b, symbolic regression with lag was conducted on two experimental scenarios. The first experiment was designed to investigate the influence of sampling frequency from models to predictions. The experiment was performed on A13, where the models were trained on 3 weeks worth of data ($28^{\text{th}}$ of August to $17^{\text{th}}$ of September in Figure \ref{figOverview}), sampled every 5, 10 and 15 minutes. All models were tested on the usual 2 week window ($18^{\text{th}}$ September to $1^{\text{st}}$ of October), sampled every 15 minutes. The results are presented in Table \ref{tabSRTrainOnAllTestOn15}. 
The second experiment was designed to investigate models' quality when used for prediction at different junctions. In this experiment we trained SL models for 3 weeks on junctions A13, A21, A36 and A51, respectively, and tested them, for 2 weeks, on the junction they were trained on, as well as all the others. The considered junctions are structurally similar but not identical: A13 has one extra input arm, therefore, to keep the experiment consistent, a lane with 0 values was added for all other junctions in both the training and testing data. The sampling interval for both training and testing is 15 minutes and the results are presented in tables \ref{tabJunctionMixRMSE} and \ref{tabJunctionMixMAE}. 

\paragraph*{Discussion} The results of the first experiment (Table \ref{tabSRTrainOnAllTestOn15}) show that the density of training data (whether samples are available every 5, 10 or 15 minutes) bares little influence on the model's accuracy over the testing window. Again, statistical training yields small p values, implying that the differences in accuracy between models trained on the three windows are significant, however, those differences are small - the highest is 1.57 cars (5 vs 15 minutes, RMSE values). As expected, the second experiment shows that testing accuracy is best on the same junction the model was trained on (the values in tables \ref{tabJunctionMixRMSE} and \ref{tabJunctionMixMAE} are smallest on the diagonals, with a slight anomaly regarding junction A36, where the model trained on A21 tests better than the ``native'' one). When models are tested on a junction different than the one used for training, the loss in accuracy peaks at 13.92 cars over 15 minutes (the difference in the RMSE scores recorded for the model trained and tested on A13 and the one trained on A51 and tested on A13). Apart from the isolated case of A51\footnote{The topography of A51 is different to that of A13, as there are two additional output branches where traffic data are not being captured. Regardless, the SL generated prediction is still reasonably accurate, albeit there is an expected drop in quality when compared against the other junctions. H-W models only consider the central junction lane, therefore their performance is not affected by other input/output lanes –- this makes them unsuitable for comparison against SR and SL models.}, the highest difference is of 5.15 cars. Our claim 2b is validated by these results, leading to the conclusion that the models generated by the symbolic regression with lag algorithm can tolerate a drop in training data density and, at the same time, may be deployed on structurally similar, yet geographically distinct junctions than the one they were trained on, without a substantial decrease in accuracy on testing data. This makes our method appealing for urban road networks, where new junctions need not necessarily be monitored, as their traffic may be predicted using an existing model obtained on a similar structure.

\begin{table*}[h!]
\begin{center}
 \caption{Symbolic regression with lag on junction A13 with varying training sampling intervals; p = $0.25 \times 10^{-16}$ %0.00000000000000256030% 
 (5 min vs 10 min); p = $0.16 \times 10^{-16}$ %0.00000000000000001633 
 (10 min vs 15 min); p = $0.43 \times 10^{-3}$ %0.00043268000000000000 
 (5 min vs 15 min)}
\begin{tabular}{c|c|c|c|c|}
& \multicolumn{2}{c|}{5 min} & \multicolumn{2}{c|}{10 min}   \\
&RMSE & MAE, StDev & RMSE & MAE, StDev   \\
&RMSE $\pm$ CI & MAE $\pm$ CI& RMSE $\pm$ CI & MAE $\pm$ CI   \\
\hline
best& 7.56 & 5.24, 5.45 & 6.85 & 4.79, 4.9  \\
 all & $8.85 \pm 0.28$ & $6.32 \pm 0.25$  & $7.46 \pm 0.10$ & $5.36 \pm 0.12$  \\
\end{tabular}

\vspace{10pt}

\begin{tabular}{c|c|c|}
& \multicolumn{2}{c|}{15 min} \\
& RMSE & MAE, StDev \\
& RMSE $\pm$ CI & MAE $\pm$ CI \\
\hline
best&  6.91 & 4.68, 5.09 \\
 all &  $7.28 \pm 0.05$ & $5.22 \pm 0.09$ \\
\end{tabular}
\label{tabSRTrainOnAllTestOn15}
\end{center}
\end{table*}

\begin{table}[h!]
\begin{center}
\caption{Symbolic regression with lag - train and test junction mix \& match; RMSE values}
\begin{tabular}{c|c|c|c|c}
\backslashbox{Trained}{Tested}  & A13 & A21 & A36 & A51\\
\hline
A13 & 6.91 & 7.71 & 9.51 & 12.60\\
\hline
A21 & 10.78 & 5.21 & 8.07 & 13.04 \\
\hline
A36 & 12.06 & 6.78 & 8.13 & 12.57\\
\hline
A51 & 20.83 & 14.42 & 21.70 & 7.52\\
\end{tabular}
\label{tabJunctionMixRMSE}
\end{center}
\end{table}

\begin{table}[h!]
\begin{center}
\caption{Symbolic regression with lag - train and test junction mix \& match; MAE values}
\begin{tabular}{c|c|c|c|c}
\backslashbox{Trained}{Tested}  & A13 & A21 & A36 & A51\\
\hline
A13 & 4.68 & 5.98 & 7.29 & 8.93 \\
\hline
A21 & 8.10 & 3.78 & 5.86 & 9.58 \\
\hline
A36 & 9.41 & 5.11 & 5.81 & 9.27\\
\hline
A51 & 17.16 & 12.16 & 17.90 & 5.38\\
\end{tabular}
\label{tabJunctionMixMAE}
\end{center}
\end{table}
%[include trees/formulae for the six junctions (or as many as possible) generated with SRwLag and point out the structural differences between junctions]
%[include statistical analysis on the trees - histograms like in EuroGP paper]

%[Junctions with data available - A13, A21, A36, A51. A13 has 5 input lanes, A21, A36 and A51 have 4 input lanes. No similar junction to A13 with sufficient data available.]

\subsection{Claim 2c: Our models are self-managing}

\begin{figure}
%[h!]
\begin{center}
\begin{subfigure}[t]{0.45\textwidth}
\includegraphics[width=\textwidth]{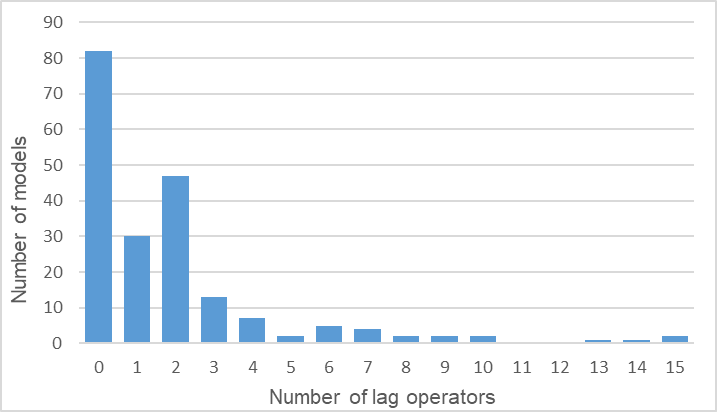}

\caption{Count of lag operators in the best 50 models for each of the junctions A13, A21, A36, A51, trained over 15 minute intervals}
\label{figFrequencyOfLag}
\end{subfigure}
\begin{subfigure}[t]{0.5\textwidth}
\includegraphics[width=\textwidth]{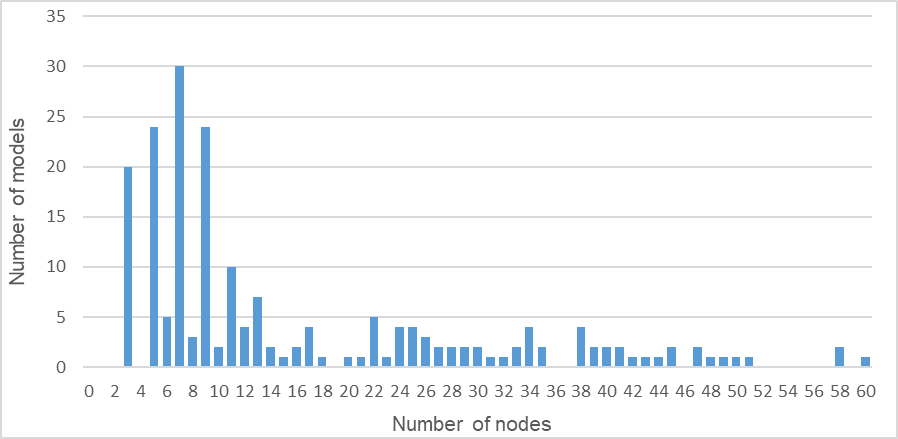}
\caption{Node count of the best 50 models for each of the junctions A13, A21, A36, A51, trained over 15 minute intervals}
\label{figTreeNodeCount}
\end{subfigure}
\caption{Statistical analysis of tree structures}
\end{center}

\end{figure}

Models for junctions A13, A21, A36 and A51 were built using symbolic regression with lag, trained on a 3 week training window ($28^{\text{th}}$ of August to $17^{\text{th}}$ of September in Fig. \ref{figOverview}), with a sampling period of 5 and 15 minutes, respectively. Tables \ref{tabAllJunctionsInputCount5min} and \ref{tabAllJunctionsInputCount15min} report the number of times each input lane, X0 through X4, is featured in the models for the four junctions. Fig.~ \ref{figFrequencyOfLag} shows the frequency of the lag operator in the best models and Fig.~ \ref{figTreeNodeCount} indicates the tree sizes of the best models.

\paragraph*{Discussion} As the results clearly show, whenever the traffic through an input lane does not significantly influence the output flow, the irrelevant junction arm will not be featured in the model. For instance, X3 bears no influence on the traffic out of A21 over the course of 5 minutes, yet this changes when the window of interest is expanded to 15 minutes and consequently X3 is featured 5 times. The elimination of irrelevant model terms is done by the SL algorithm in an unsupervised way. 
The models evolve to contain the necessary elements (lag can occur between zero and 15 times in a model, as shown in the histogram of Fig. \ref{figFrequencyOfLag}),  and to be of a size that allows for good predictions (as indicated in the histogram of Fig. \ref{figTreeNodeCount}, varying between 3 and 60 nodes), thus validating claim 2c.
Furthermore, the data in Tables \ref{tabAllJunctionsInputCount5min} and \ref{tabAllJunctionsInputCount15min} represent occurrences of different input as opposed to linear regression, which would show weights. The advantage of SL is that the granularity of the model adapts to the data. As the sampling interval changes, the models reflect this by featuring both the relevant input and the relevant operators as often as needed. In other words, the nonlinear models become more complex or more simple in direct response to traffic conditions.

% \begin{table}[h!]
% \begin{tabular}{c|c|c|c|c|c}
%  & X0 & X1 & X2 & X3 & X4 \\
%  5min & 59 & 67 & 25 & 1 & 3 \\
% 15min & 71 & 71 & 53 & 10 & 5 \\
% \end{tabular}
% \caption{Junction A13 count of inputs used in trees for 5 and 15 minute intervals.}
% \label{tabA13InputCount}
% \end{table}

\begin{table}[t!]
\begin{center}
\caption{Symbolic regression with lag on all junctions - input count in models trained with 5 minute sampling}
\begin{tabular}{c|c|c|c|c|c}
 & X0 & X1 & X2 & X3 & X4 \\
 \hline
 A13 & 59 & 67 & 25 & 1 & 3 \\
A21 & 26 & 12 & 129 & 0  & n/a\\
A36 & 85 & 3 & 51 & 70& n/a \\
A51 & 0 & 79 & 24 & 6& n/a \\
\end{tabular}
\label{tabAllJunctionsInputCount5min}
\end{center}
\end{table}

\begin{table}[h!]
\begin{center}
\caption{Symbolic regression with lag on all junctions - input count in models trained with 15 minute sampling}
\begin{tabular}{c|c|c|c|c|c}
 & X0 & X1 & X2 & X3 & X4 \\
 \hline
A13 & 71 & 71 & 53 & 10 & 5 \\
A21 & 41 & 14 & 142 & 5 & n/a\\
A36 & 72 & 6 & 29 & 83 & n/a \\
A51 & 11 & 74 & 13 & 32& n/a \\
\end{tabular}
\label{tabAllJunctionsInputCount15min}
\end{center}
\end{table}

\subsection{Claim 3a: Our models have a long shelf life}
Symbolic regression with lag was trained on a 3 week training window, spanning from the 28\textsuperscript{th} of August to the 17\textsuperscript{th} of September. Testing was performed on two time intervals: the first is the usual window (18\textsuperscript{th} of September to the 1\textsuperscript{st} of October), immediately after the training period, and the second spans from 20\textsuperscript{th} November 2017 to 3\textsuperscript{rd} of December 2017, which is 9 weeks after the training period. Both experiments were conducted on junction A13, with a 15 minute sampling interval, for both training and testing. 

\paragraph*{Discussion} In the case where training was performed at the end of August and testing at the beginning of December, the RMSE was 7.87, less than one car higher, compared to the RMSE of 6.91 for the case where the training and testing periods make a continuous block of time.
 Thus, our claim 3a is validated, in the sense that the accuracy of models produced by SL does not decay substantially over time. This is particularly useful from a computational standpoint, as traffic engineers and  urban planners need not retrain older models in order to accurately predict traffic at a later date. 
%\begin{table}[h!]
%\caption{Symbolic regression with lag on junction A13 - tested straight after training (left) and 9 weeks later (right)}
%\begin{tabular}{c|c|c|c|c|}
%& \multicolumn{2}{c|}{Straight after} & \multicolumn{2}{c|}{9 weeks later} \\
%& RMSE & MAE, StDev & RMSE & MAE, StDev \\
%15 min& 6.91 & 4.68, 5.09 & 7.87 & 5.81, 5.31  \\
%\end{tabular}
%\label{tabLSTestedLater}
%\end{table}

\begin{figure}[h!]
\begin{center}
\includegraphics[width=0.75\textwidth]{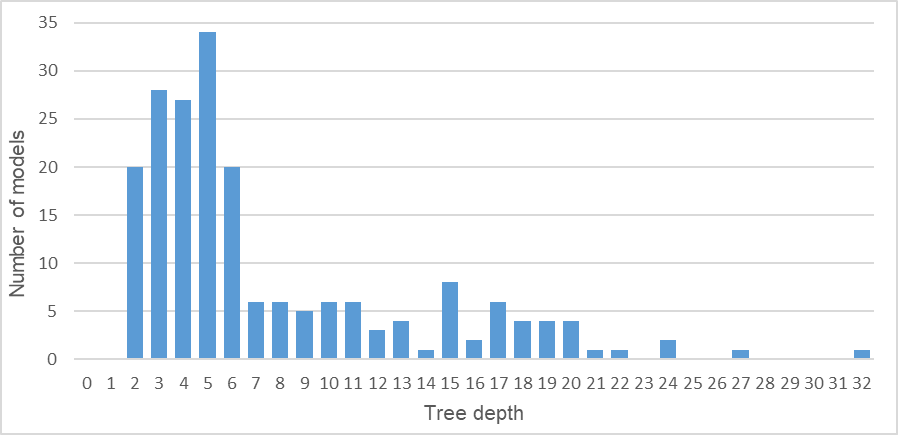}
\caption{Depth of trees of the best 50 models for each of the junctions A13, A21, A36, A51, trained over 15 minute intervals}
\label{figTreeDepth}
\end{center}
\end{figure}

\subsection{Claim 3b: Our models are easy to understand, suitable for effective and confident decision making}

Symbolic regression with lag was run on 3 weeks of training data ($28^{\text{th}}$ of August to $17^{\text{th}}$ of September in Fig. \ref{figA13MissingData}), sampled every 5 and 15 minutes, respectively. The experiments were conducted on junctions A13, A21, A36 and A51. Figure \ref{figTreeDepth} reports the number of trees of depths 0 through 32, over the set of 50 best models obtained from as many runs, for each junction.
\paragraph*{Discussion} Most of the generated trees, namely 129 out of 200, for 15 minute training intervals, have depths between 2 and 6 levels, a manageable size, rendering the respective models easy to inspect by the human eye. The largest tree, produced only once over all experiments (for A13, for a sampling interval of 15 minutes), is 32 levels deep and contains 49 nodes, which is still acceptable for a human stakeholder seeking to extract infrastructure relevant value from it. Of course, the operator may always choose a simpler tree from the range available (see Fig. \ref{figTreeDepth}). Given the manageable size of the models produced by the proposed symbolic regression with lag algorithm, as well as their clear structure, where model terms can be easily mapped against physical junction lanes (See Fig. \ref{figA13_SymbRegressionWithLag}), our claim 3b is verified.

\section{Conclusions}
\label{secConclusions}
%[restate the problem and our proposed solution- provide a practical implementation of ITS using a new approach based on symbolic regression with a lag operator]
Life in major urban communities is heavily impacted by the quality of road traffic. Intelligent Transportation Systems are meant to alleviate the negative effects of the ever increasing vehicle flow through our cities, by decongesting busy junctions, reducing maintenance costs and cutting down commute times. This is done with the aid of advanced computing techniques, such as data mining and forecasting, employed in the processes of monitoring, modelling and predicting traffic behaviour. The end goal is to extract insight from vast amounts of collected traffic data, in a form that is suitable for decision support, i.e., informing traffic authorities with regard to likely traffic patterns through a junction yet to be built. 

%[summarise main contributions in 2 sentences, explaining how we are better than the state of the art]
In this paper, we contribute a new forecasting technique, based on symbolic regression enhanced with a lag operator. The performance of the proposed method is compared against time series approaches, namely Holt-Winters triple exponential smoothing, as well as classic linear regression and simple symbolic regression, all applied on the Darmstadt city road network. The detailed experimental analysis supports our claims. Specifically, symbolic regression with lag produces models that
\begin{itemize}
\item feature a high tolerance to missing data samples, 
\item are robust and reliable even when trained on a time window as short as one week,
\item are sufficiently flexible to model urban traffic, as opposed to the simpler case of highways,
\item self-manage, in the sense that models adapt to real traffic conditions,
\item have a long shelf-life, in that they remain valid at least up to 9 weeks after they were trained,
\item are easy to understand, and thus suitable to support urban planning decision making.
\end{itemize}
In the immediate future, we welcome the possibility that a major European city would test out their urban infrastructure plans by using our approach, before implementing it.
\section*{Acknowledgement}
This work was supported by the European Commission through the H2020 project EXCELL (\url{http://excell-project.eu/}), grant No. 691829.
\newpage
\section*{References}

\bibliography{main}

\end{document}